\RequirePackage[2014/01/01]{latexrelease}
\documentclass[10pt,journal,compsoc]{IEEEtran}
\usepackage{graphicx}
\usepackage{multirow}
\usepackage{amsmath,amssymb} 
\usepackage{color}
\usepackage[table,xcdraw]{xcolor}
\usepackage[outercaption]{sidecap}
\usepackage{floatrow}
\usepackage{adjustbox}
\usepackage{url}
\usepackage{hyperref}

\ifCLASSOPTIONcompsoc
  \usepackage[nocompress]{cite}
\else
  \usepackage{cite}
\fi

\usepackage[outerbars]{changebar}
\def\etal{\emph{et al. }}

\begin{document}

\bstctlcite{IEEEexample:BSTcontrol}

\title{A Systematic Evaluation and Benchmark for Person Re-Identification: Features, Metrics, and Datasets}

\author{Srikrishna~Karanam$^{*}$,~\IEEEmembership{Student~Member,~IEEE,}
        Mengran~Gou$^{*}$,~\IEEEmembership{Student~Member,~IEEE,}
        Ziyan~Wu,~\IEEEmembership{Member,~IEEE,}
        Angels Rates-Borras,
        Octavia~Camps,~\IEEEmembership{Member,~IEEE,}
        and~Richard~J.~Radke,~\IEEEmembership{Senior~Member,~IEEE}
\IEEEcompsocitemizethanks{\IEEEcompsocthanksitem  R.J. Radke is with the Department of Electrical, Computer, and Systems Engineering, Rensselaer Polytechnic Institute, Troy, NY, 12180 (e-mail: rjradke@ecse.rpi.edu).\vspace{-1em}}
\IEEEcompsocitemizethanks{\IEEEcompsocthanksitem S. Karanam and Z. Wu are with Siemens Corporate Technology, Princeton, NJ 08540 (e-mail: srikrishna.karanam@siemens.com, ziyan.wu@siemens.com).\vspace{-1em}}
\IEEEcompsocitemizethanks{\IEEEcompsocthanksitem  M. Gou, A. Rates-Borras, and O. Camps are with the Department of Electrical and Computer Engineering, Northeastern University, Boston, MA, 02115 (e-mail: mengran@coe.neu.edu, ratesborras.a@husky.neu.edu, camps@coe.neu.edu).\vspace{-1em}}
\IEEEcompsocitemizethanks{\IEEEcompsocthanksitem $^{*}$S.~Karanam and M.~Gou contributed equally to this work. Corresponding author: S.~Karanam.}
}

\IEEEtitleabstractindextext{
\begin{abstract}
Person re-identification (re-id) is a critical problem in video analytics applications such as security and surveillance. The public release of several datasets and code for vision algorithms has facilitated rapid progress in this area over the last few years. However, directly comparing re-id algorithms reported in the literature has become difficult since a wide variety of features, experimental protocols, and evaluation metrics are employed. In order to address this need, we present an extensive review and performance evaluation of single- and multi-shot re-id algorithms.  The experimental protocol incorporates the most recent advances in both feature extraction and metric learning. To ensure a fair comparison, all of the approaches were implemented using a unified code library that includes 11 feature extraction algorithms and 22 metric learning and ranking techniques. All approaches were evaluated using a new large-scale dataset that closely mimics a real-world problem setting, in addition to 16 other publicly available datasets: VIPeR, GRID, CAVIAR, DukeMTMC4ReID, 3DPeS, PRID, V47, WARD, SAIVT-SoftBio, CUHK01, CHUK02, CUHK03, RAiD, iLIDSVID, HDA+, and Market1501. The evaluation codebase and results will be made publicly available for community use. 
\end{abstract}
}

\maketitle

\IEEEpeerreviewmaketitle

\section{Introduction}
\label{sec:introduction}

\IEEEPARstart{P}{erson} re-identification, or re-id, is a critical task in most surveillance and security applications \cite{li2014real,gong2014person,camps2016from} and has increasingly attracted attention from the computer vision community \cite{gray2008viewpoint,prosser2010person,zheng2011person,koestinger2012large,mignon2012pcca,zhao2013unsupervised,bazzani2013symmetry,pedagadi2013local,an2013reference,xiong2014person,zhao2014learning,wu2015viewpoint,paisitkriangkrai2015learning,karanam2015person,liao2015person,zheng2015partial,li2015multiscale,messelodi2015boosting,chen2015similarity,gou2016person}.  The fundamental re-id problem is to compare a person of interest as seen in a ``probe'' camera view to a ``gallery" of candidates captured from a camera that does not overlap with the probe one.  If a true match to the probe exists in the gallery, it should have a high matching score, or rank, compared to incorrect candidates.

Since the body of research in re-id is now quite large, we can begin to draw conclusions about the best combinations of algorithmic subcomponents.  In this paper, we present a careful, fair, and systematic evaluation of feature extraction, metric learning, and multi-shot ranking algorithms proposed for re-id on a wide variety of benchmark datasets. Our general evaluation framework is to consider all possible combinations of feature extraction and metric learning algorithms for single-shot datasets and all possible combinations of feature extraction, metric learning, and multi-shot ranking algorithms for multi-shot datasets. In particular, we evaluate 276 such algorithm combinations on 10 single-shot re-id datasets and 646 such algorithm combinations on 7 multi-shot re-id datasets, making the proposed study the \textbf{largest and most systematic} re-id benchmark to date. As part of the evaluation, we built a \textbf{public code library} with an easy-to-use input/output code structure and uniform algorithm parameters that includes 11 contemporary feature extraction and 22 metric learning and ranking algorithms. Both the code library and the complete benchmark results are publicly available for community use at {\color{blue}\url{https://github.com/RSL-NEU/person-reid-benchmark}}.

\footnotetext[1]{\scriptsize This material is based upon work supported by the U.S. Department
of Homeland Security under Award Number 2013-ST-061-ED0001. The views and conclusions contained in this document are those of the authors and should not be interpreted as necessarily representing the official policies, either expressed or implied, of the U.S. Department of Homeland Security. Thanks to Michael Young, Jim Spriggs, and Don Kemer for supplying the airport video data.}

\begin{figure*}[t]
\centering
\includegraphics[width=0.6\textwidth]{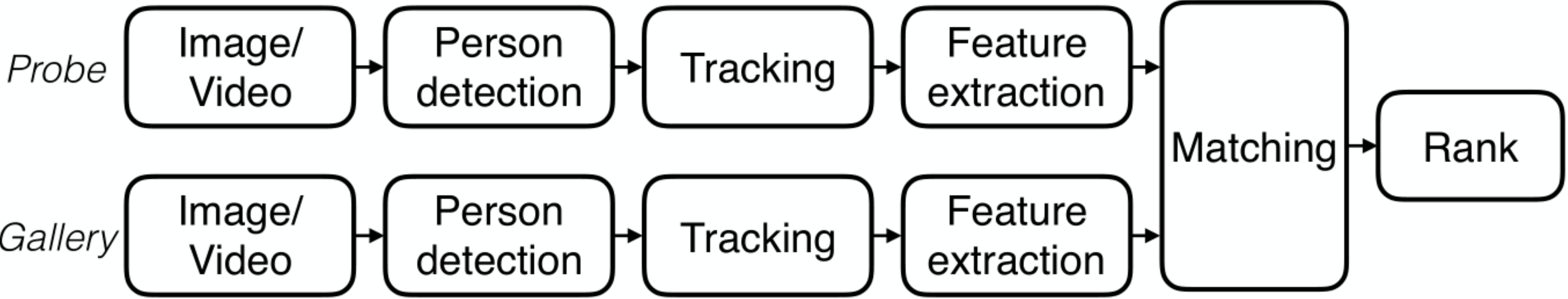}
\caption{A typical end-to-end re-id system pipeline.}
\label{fig:flowchart}
\end{figure*}

Existing re-id algorithms are typically evaluated on academic re-id datasets \cite{gray2008viewpoint,cheng2011custom,hirzer2011person,baltieri20113dpes,martinel2012distributed,wang2014person,das2014consistent,loy2013person} that are specifically hand-curated to only have sets of bounding boxes for the probes and the corresponding matching candidates.  On the other hand, real-world end-to-end surveillance systems include automatic detection and tracking modules, depicted in Figure~\ref{fig:flowchart}, that generate candidates on-the-fly, resulting in gallery sets that are dynamic in nature. Furthermore, errors in these modules may result in bounding boxes that may not accurately represent a human \cite{camps2016from}. While these issues are critical in practical re-id applications, they are not well-represented in the currently available datasets. To this end, our evaluation also includes a \textbf{new, large-scale dataset} constructed from images captured in a challenging surveillance camera network from an airport. All the images in this dataset were generated by running a prototype end-to-end real-time re-id system using automatic person detection and tracking algorithms instead of hand-curated bounding boxes.   

\section{Evaluated Techniques}
\label{sec:bench}
In this section, we summarize the feature extraction, metric learning, and multi-shot ranking techniques that are evaluated as part of the proposed re-id benchmark, which include algorithms published through ECCV 2016. We anticipate that the benchmark will be updated (along the lines of the Middlebury benchmarks \cite{scharstein2002taxonomy,baker2011database}) as new techniques are implemented into our evaluation framework.

\subsection{Feature extraction}
\label{sec:bench:feature}
We consider 11 feature extraction schemes that are commonly used in the re-id literature, summarized in Table \ref{tab:featMetric}(a).  In ELF \cite{gray2008viewpoint}, color histograms in the RGB, YCbCr, and HS color spaces, and texture histograms of responses of rotationally invariant Schmid \cite{schmid2001constructing} and Gabor \cite{fogel1989gabor} filters are computed. In LDFV \cite{ma2012local}, local pixel descriptors comprising pixel spatial location, intensity, and gradient information are encoded into the Fisher vector \cite{sanchez2013image} representation. In gBiCov \cite{ma2014covariance}, multi-scale biologically-inspired features \cite{riesenhuber1999hierarchical} are encoded using covariance descriptors \cite{tuzel2008pedestrian}. In  IDE-CaffeNet, IDE-ResNet, and IDE-VGGNet, we use the idea first presented in the DeepFace paper \cite{taigman2014deepface} and applied to re-id by Zheng \etal \cite{zheng2016mars}, in which every person is treated as a separate class and a convolutional neural network is trained for a classification objective. AlexNet \cite{krizhevsky2012imagenet}, ResNet \cite{he2015deep}, and VGGNet \cite{simonyan2014very} architectures are employed in IDE-CaffeNet, IDE-ResNet and IDE-VGGNet respectively. In each case, we start with a model pre-trained on the ImageNet dataset, and finetune it using the datasets we consider in this evaluation. Specifically, during finetuning, we modify the weights of all the layers of the network. More implementation details are presented in Section~\ref{sec:bench:eval}.  In DenseColorSIFT \cite{zhao2013unsupervised}, each image is densely divided into patches, and color histograms and SIFT features are extracted from each patch. In HistLBP \cite{xiong2014person}, color histograms in the RGB, YCbCr, and HS color spaces and texture histograms from local binary patterns (LBP) \cite{ojala1996comparative} features are computed. In LOMO \cite{liao2015person}, HSV color histograms and scale-invariant LBP \cite{liao2010modeling} features are extracted from the image processed by a multi-scale Retinex algorithm \cite{jobson1997multiscale}, and maximally-pooled along the same horizontal strip. In GOG \cite{matsukawa2016hierarchical}, an image is divided into horizontal strips and local patches in each strip are modeled using a Gaussian distribution. Each strip is then regarded as a set of such Gaussian distributions, which is then summarized using a single Gaussian distribution. 

\begin{table*}[htb!]
\caption{Evaluated feature extraction and metric learning methods.}
\renewcommand{\arraystretch}{1.2}
\centering
\begin{tabular}{cc}
\begin{tabular}[t]{l|l}
Feature &Year\\
\hline
ELF \cite{gray2008viewpoint} & ECCV 08 \\ \hline
LDFV \cite{ma2012local} & ECCVW 12 \\ \hline
gBiCov \cite{ma2014covariance} & BMVC 12 \\ \hline
IDE-CaffeNet \cite{krizhevsky2012imagenet,zheng2016person} & NIPS 12, ECCV 16 \\ \hline
DenseColorSIFT \cite{zhao2013unsupervised} & CVPR 13 \\ \hline
HistLBP \cite{xiong2014person} & ECCV 14 \\ \hline
IDE-VGGNet \cite{simonyan2014very,zheng2016person} & ICLR 15, ECCV 16 \\ \hline
LOMO \cite{liao2015person} & CVPR 15 \\ \hline
IDE-ResNet \cite{he2015deep,zheng2016person} & ICCV 15, ECCV 16 \\ \hline
WHOS \cite{lisanti2015person} & T-PAMI 15 \\ \hline
GOG \cite{matsukawa2016hierarchical} & CVPR 16
\end{tabular}&
\begin{tabular}[t]{l|l||l|l}
Metric & Year & Metric & Year\\
\hline
$l_{2}$ & & kPCCA \cite{mignon2012pcca} & CVPR 12 \\ \hline
FDA \cite{fisher1936use} & AE 1936 & LFDA \cite{pedagadi2013local} & CVPR 13 \\  \hline 
ITML \cite{davis2007information} & ICML 07 & SVMML \cite{li2013learning} & CVPR 13\\ \hline
MFA \cite{yan2007graph} & PAMI 07 & kMFA\cite{xiong2014person}  & CVPR 13 \\ \hline
LMNN \cite{weinberger2009distance} & JMLR 08 & KCCA \cite{lisanti_icdsc14} & ICDSC 14  \\ \hline
RankSVM \cite{prosser2010person} & BMVC 10 & rPCCA \cite{xiong2014person} & ECCV 14 \\ \hline 
PRDC \cite{zheng2011person} & CVPR 11 & kLFDA \cite{xiong2014person} & ECCV 14 \\ \hline
KISSME \cite{koestinger2012large} & CVPR 12 & XQDA \cite{liao2015person} & CVPR 15 \\ \hline
PCCA \cite{mignon2012pcca} & CVPR 12 & NFST \cite{zhang2016learningnull} & CVPR 16 
\end{tabular} \\
(a) & (b)
\end{tabular}
\label{tab:featMetric}
\vspace*{-2em}
\end{table*}

\subsection{Metric learning}
\label{sec:bench:metric}
While using any of the features described in the previous section in combination with the Euclidean distance (\texttt{$l_{2}$}) can be used to rank gallery candidates, this would be an unsupervised and suboptimal approach. Incorporating supervision using training data leads to superior performance, which is the goal of metric learning, i.e., learning a new feature space such that feature vectors of the same person are close whereas those of different people are relatively far. We consider 18 metric learning methods that are typically used by the re-id community, summarized in Table \ref{tab:featMetric}(b).  Fisher discriminant analysis (FDA) \cite{fisher1936use}, local Fisher discriminant analysis (LFDA) \cite{pedagadi2013local}, marginal Fisher analysis (MFA) \cite{yan2007graph}, cross-view quadratic discriminant analysis (XQDA) \cite{liao2015person}, and discriminative null space learning (NFST) \cite{zhang2016learningnull} all formulate a Fisher-type optimization problem that seeks to minimize the within-class data scatter while maximizing between-class data scatter. In practice, scatter matrices are regularized by a small fraction of their trace to deal with matrix singularities. Information-theoretic metric learning (ITML) \cite{davis2007information}, large-margin nearest neighbor (LMNN) \cite{weinberger2009distance}, relative distance comparison (PRDC) \cite{zheng2011person}, keep-it-simple-and-straightforward metric (KISSME) \cite{koestinger2012large}, and pairwise constrained component analysis (PCCA) \cite{mignon2012pcca} all learn Mahalanobis-type distance functions using variants of the basic pairwise constraints principle. kPCCA \cite{mignon2012pcca}, kLFDA \cite{xiong2014person}, and kMFA \cite{xiong2014person} kernelize PCCA, LFDA, and MFA, respectively. kCCA \cite{lisanti_icdsc14} adopts canonical correlation analysis to map the kernelized features into a common subspace. For these kernel-based methods, we consider the standard linear, exponential (exp), chi2 (${\chi^2}$), and chi2-rbf (\texttt{R$_{\chi^2}$}) kernels. In RankSVM \cite{prosser2010person}, a weight vector that weights the different features appropriately is learned using a soft-margin SVM formulation. In SVMML \cite{li2013learning}, locally adaptive decision functions are learned in a large-margin SVM framework.

\subsection{Multi-shot ranking}
\label{sec:bench:multi}
While most re-id algorithms are single-shot, i.e.,  features are extracted from a single probe image of the person of interest, the multi-shot scenario, in which features are extracted from a series of images of the person of interest, is arguably more relevant to video analysis problems.  The simplest way to handle multi-shot data is to compute the average feature vector for each person, effectively resulting in a single-shot problem.  However, we also evaluated several algorithms that inherently address multi-shot data, treating it as an image set and constructing affine hulls to compute the distance between a gallery and a probe person. Specifically, we considered the AHISD \cite{cevikalp2010face} and RNP \cite{yang2013face} algorithms.  While these methods were proposed in the context of face recognition, the basic notion of image set matching applies to re-id as well. We also evaluated a multi-shot method based on sparse ranking, in which re-id is posed as a sparse recovery problem. Specifically, we consider SRID \cite{karanam2015sparse}, where a block sparse recovery problem is solved to retrieve the identity of a probe person, and ISR \cite{lisanti2015person}, where the recovered sparse coefficient vector is re-weighted using an iterative scheme to rank gallery candidates and re-identify the person of interest. 

\subsection{Techniques not (yet) considered}
\label{sec:bench:misc}
As noted in Section~\ref{sec:introduction}, the framework we adopt involves evaluating all possible combinations of candidate feature extraction, metric learning, and multi-shot ranking algorithms. 
Methods that do not fall into this evaluation framework include post-rank learning methods \cite{liu2013pop,garcia2015person}, unsupervised learning \cite{zhao2013unsupervised,cheng2011custom,bak2011multiple}, attribute learning \cite{su2015multi,layne2012person,shi2015transferring}, ensemble methods \cite{eisenbach2015evaluation,zheng2015query,martinel2016pool} and mid-level representation learning \cite{zhao2014learning}. A more comprehensive survey of these and other related methods can be found in the book by Gong \etal \cite{gong2014person} and papers by Zheng \cite{zheng2016person}, Satta \cite{satta2013appearance}, Vezzani \cite{vezzani2013people}, and Bedagkar-Gala and Shah \cite{bedagkar2014survey}.  While these methods are currently not part of our evaluation, we plan to expand our study and include them in a future release. We note that while all the evaluated algorithms follow a two-step process of feature and metric learning, we do consider a baseline Siamese CNN \cite{ahmed2015improved} algorithm that learns features and metrics together in a single-step approach.  While we provide extensive discussion about challenges and opportunities in learning more powerful architectures that follow this single-step approach in Sections~\ref{sec:bench:results:misc:datasets} and Sec~\ref{sec:futureresearch}, our results suggest that the two-step approach also gives competitive performance, and more importantly, can provide re-id specific domain knowledge and insights to aid future research.

\section{Datasets}
\label{sec:bench:datasets}
In this section, we briefly summarize the various publicly available datasets that are used in our benchmark evaluation. Table~\ref{tab:dataset} provides a statistical summary of each dataset.  Based on difficult examples, we also annotate each dataset with challenging attributes from the following list: viewpoint variations (VV), illumination variations (IV), detection errors (DE), occlusions (OCC), background clutter (BC), and low-resolution images (RES).  We also indicate the number of bounding boxes (BBox), false positives (FP), distractors (distrac.) and cameras (cam) in each dataset, and the means by which the bounding boxes were obtained: using hand-labeling (hand), aggregated channel features \cite{dollar2014fast} (ACF), or the deformable parts model detector \cite{felzenszwalb2010object} (DPM). Samples of  difficult examples are provided as part of the supplementary material \footnote{Supplementary material can be found at {\color{blue}\url{https://arxiv.org/abs/1605.09653}}.}.

\begin{table*}[htb!]
\setlength{\tabcolsep}{11pt}
\renewcommand{\arraystretch}{1.05}
\centering
\caption{The characteristics of the 17 datasets of the re-id benchmark.}
\label{tab:dataset}
\begin{tabular}{ccccccccc}
\hline
Dataset      		& \# people & \# BBox   &\# FP+distrac. & \# cam	& label  & Attributes  \\ \hline
VIPeR        		& 632           & 1,264    	& 0             & 2         & hand   &  VV,IV \\
\rowcolor[HTML]{EFEFEF} 
GRID         		& 1025           & 1,275    	& 775           & 2         & hand   & VV,BC,OCC,RES \\ 
CAVIAR       		& 72            & 1,220    	& 0             & 2 	   	& hand   & VV,RES \\ 
\rowcolor[HTML]{EFEFEF} 
3DPeS         		& 192           & 1,011    	& 0             & 8         & hand   & VV,IV \\ 
PRID          		& 178           & 24,541    	& 0             & 2         & hand   & VV,IV \\ 
\rowcolor[HTML]{EFEFEF} 
V47           		& 47            &  752       	& 0             & 2         & hand   & - \\ 
WARD          		& 70            & 4,786    	& 0             & 3         & hand    & IV \\ 
\rowcolor[HTML]{EFEFEF} 
SAIVT-Softbio 		& 152           & 64,472    	& 0             & 8         & hand   & VV,IV,BC \\ 
CUHK01				& 971			& 3,884		& 0 			& 2 		& hand 		&VV,OCC	 \\
\rowcolor[HTML]{EFEFEF} 
CUHK02				& 1,816			& 7,264	 	& 0 			& 10		& hand 		&VV,OCC   	\\
CUHK03       		& 1,360         & 13,164   	& 0             & 10         & DPM/hand  &VV,DE,OCC\\ 
\rowcolor[HTML]{EFEFEF} 
RAiD      			& 43            & 6,920    	& 0             & 4         & hand    & VV,IV\\ 
iLIDSVID      		& 300           & 42,495   	& 0             & 2         & hand   & VV,IV,BC,OCC \\ 
\rowcolor[HTML]{EFEFEF} 
HDA+          		& 53            & 2,976     & 0             & 12        & ACF/hand & VV,IV,DE\\ 
Market1501 			& 1,501      	& 32,217   	 & 2,793+500k     & 6        	& DPM    & VV,DE,RES \\ 
\rowcolor[HTML]{EFEFEF} 
DukeMTMC4ReID 		& 1,852			& 46,261 & 21,990		& 8			& Doppia & VV,IV,DE,BC,OCC \\
Airport       		& 9,651         & 39,902    & 17,928       	& 6      	& ACF    & VV,IV,DE,BC,OCC \\ \hline
\end{tabular}
\vspace*{-1em}
\end{table*}

VIPeR \cite{gray2008viewpoint} consists of 632 people from two disjoint views. Each person has only one image per view. VIPeR suffers from subtantial viewpoint and illumination variations. GRID \cite{loy2010time} has 250 image pairs collected from 8 non-overlapping cameras. To mimic a realistic scenario, 775 non-paired people are included in the gallery set, which makes this dataset extremely challenging. GRID suffers from viewpoint variations, background clutter, occlusions and low-resolution images. CAVIAR \cite{cheng2011custom} is constructed from two cameras in a shopping mall. Of the 72 people, we only use 50 people who have images from both cameras. CAVIAR suffers from viewpoint variations and low-resolution images. In the case of 3DPeS \cite{baltieri20113dpes}, the re-id community uses a set of selected snapshots instead of the original video, which includes 192 people and 1,011 images. 3DPeS suffers from viewpoint and illumination variations. PRID \cite{hirzer2011person} is constructed from two outdoor cameras, with 385 tracking sequences from one camera and 749 tracking sequences from the other camera. Among them, only 200 people appear in both views. To be consistent with previous work \cite{wang2014person}, we use the same subset of the data with 178 people. PRID suffers from viewpoint and illumination variations. V47 \cite{wang2011re} contains 47 people walking through two indoor cameras. WARD \cite{martinel2012distributed} collects 4,786 images of 70 people in 3 disjoint cameras. SAIVT-Softbio \cite{bialkowski2012database} consists of 152 people as seen from a surveillance camera network with 8 cameras. To be consistent with existing work \cite{bialkowski2012database}, we use only two camera pairs: camera 3 and camera 8 (which we name SAIVT-38) and camera 5 and camera 8 (which we name SAIVT-58). Both these datasets suffer from viewpoint and illumination variations, and background clutter. CUHK01 \cite{li2012human} has 971 people and 3,884 images captured from 2 disjoint camera views in a college campus setting. CUHK02 \cite{li2013locally} has 1816 people and 7,264 images captured from 5 disjoint camera pairs in a college campus. All bounding boxes are manually labeled. CUHK03 \cite{li2014deepreid} has 1360 people and 13,164 images from 5 disjoint camera pairs. Both manually labeled bounding boxes and automatically detected bounding boxes using the DPM detector \cite{felzenszwalb2010object} are provided. We only use the detected bounding boxes in our experiments. CUHK03 suffers from viewpoint variations, detection errors, and occlusions. RAiD \cite{das2014consistent} includes 43 people as seen from two indoor and two outdoor cameras and suffers from viewpoint and illumination variations. iLIDSVID \cite{wang2014person} includes  600 tracking sequences for 300 people from 2 non-overlapping cameras in an airport and suffers from viewpoint and illumination variations, background clutter, and occlusions. HDA+ \cite{figueira2014hda+} was proposed to be a testbed for an automatic re-id system. Fully labeled frames for 30-minute long videos from 13 disjoint cameras are provided. Since we only focus on the re-id problem, we use pre-detected bounding boxes generated using the ACF \cite{dollar2014fast} detector. DukeMTMC4ReID \cite{gou2017duke} has 1852 identities with 46,261 images and 21,551 false alarms from the Doppia person detector \cite{benenson2014ten}. The images are captured from a disjoint 8-camera network located at the Duke University campus. This dataset was constructed specifically for re-id from the DukeMTMC multi-camera multi-target tracking dataset \cite{ristani2016performance}. Market1501 \cite{zheng2015scalable} has 1,501 people with 32,643 images and 2,793 false alarms from the DPM person detector \cite{felzenszwalb2010object}. Besides these, an additional 500,000 false alarms and non-paired people are also provided to emphasize practical problems in re-id. Market1501 suffers from viewpoint variations, detection errors and low-resolution. Airport is the new dataset we introduce in the next section.

\subsection{A new, real-world, large-scale dataset}
\label{sec:newDataset}
In this section, we provide details about a new re-id dataset we designed for this benchmark. The dataset was created using videos from six cameras of an indoor surveillance network in a mid-sized airport; this testbed is described further in \cite{camps2016from}.  The cameras cover various parts of a central security checkpoint area and three concourses.  Each of our cameras has $768\times 432$ pixels and captures video at 30 frames per second. 12-hour long videos from 8 AM to 8 PM were collected from each of these cameras.  Under the assumption that each target person takes a limited amount of time to travel through our camera network, each of these long videos was randomly split into 40 five minute long video clips. Each video clip was then run through a prototype end-to-end re-id system comprised of automatic person detection and tracking algorithms. Specifically, we employed the ACF framework of Dollar \etal \cite{dollar2014fast} to detect people and a combination of FAST corner features \cite{rosten2010faster} and the KLT tracker \cite{lucas1981iterative} to track people and associate any broken ``tracklets''. 
The dataset can be requested at {\color{blue} \href{http://www.northeastern.edu/alert/transitioning-technology/alert-datasets/alert-airport-re-identification-dataset/}{http://www.northeastern.edu/alert/transitioning-technology/alert-datasets/alert-airport-re-identification-dataset/}}.  

Unlike other datasets that capture image data from public environments such as universities \cite{gou2017duke,ristani2016performance} \cite{zheng2016mars} \cite{zhao2014learning}, shopping locations \cite{cheng2011custom}, or publicly accessible spots in transportation gateways \cite{loy2013person}, the Airport dataset provides data captured from video streams inside the secure area, post the security checkpoint, of a major airport. It is generally very difficult to obtain data from such a camera network, in which configuration settings (e.g., network topology and placement of cameras) are driven by security requirements. For instance, most academic datasets summarized in Table \ref{tab:dataset} have images taken from cameras with optical axes parallel to the ground plane, as opposed to the real world where the angle is usually much larger due to constraints on where and how the cameras can be installed. This aspect is explicitly captured by the Airport dataset. Unlike other datasets that primarily capture images of people in a university setup (e.g., Market1501, CUHK, DukeMTMC4ReID), the Airport dataset captures images of people from an eclectic mix of professions, leading to a richer, more diversified set of images. Another key difference with existing datasets is the temporal aspect; we capture richer time-varying crowd dynamics, i.e., the density of people appearing in the source videos naturally varies according to the flight schedule at each hour.  Such time-varying behavior can help evaluate the temporal performance of re-id algorithms, an understudied area \cite{karanam2017rank}.  

\begin{figure*}[t]
\centering
\includegraphics[scale=0.5]{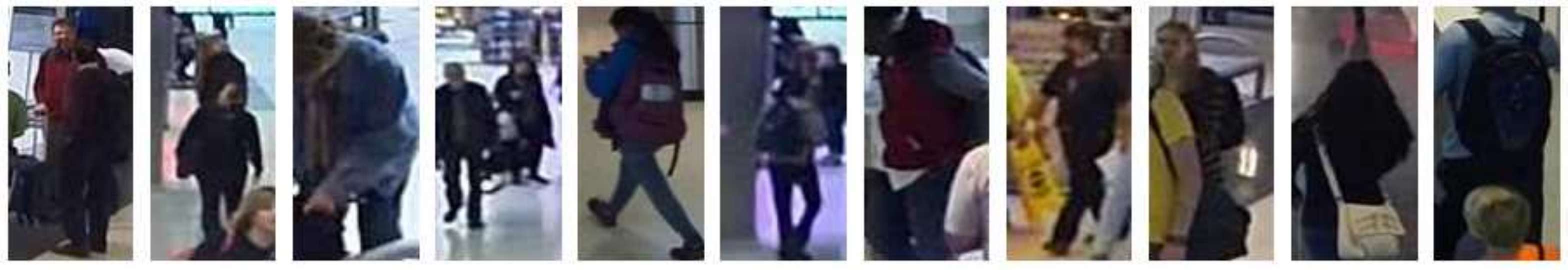}
\caption{Samples of images from the proposed Airport dataset. See supplementary material for more snapshots.}
\label{fig:sampleImages}
\end{figure*}

Since all the bounding boxes were generated automatically without any manual annotation, this dataset accurately mimics a real-world re-id problem setting. A typical fully automatic re-id system should be able to automatically detect, track, and match people seen in the gallery camera, and the proposed dataset exactly reflects this setup. In total, from all the short video clips, tracks corresponding to 9,651 unique people were extracted. The number of bounding box images in the dataset is 39,902, giving an average of 3.13 images per person. The sizes of detected bounding boxes range from 130$\times$54 to 403$\times$166. 1,382 of the 9,651 people are paired in at least two cameras. A number of unpaired people are also included in the dataset to simulate how a real-world re-id system would work: given a person of interest in the probe camera, a real system would automatically detect and track all the people seen in the gallery camera. Therefore, having a dataset with a large number of unpaired people greatly facilitates algorithmic re-id research by closely simulating a real-world environment. While this aspect is discussed in more detail in our system paper \cite{camps2016from}, we briefly describe how this dataset can be used to validate detection and tracking algorithms typically used in an end-to-end re-id system. Specifically, since we have both valid and invalid detections in our dataset, we can use them interchangeably to evaluate the impact of the detection module. For instance, adding invalid detections to the gallery would help evaluate the need for more detection accuracy at the cost of compute time. Since we have access to multiple broken tracklets for each person, we can interchangeably use them to evaluate the impact of the tracking module. For instance, manually associating all broken tracklets can help evaluate the need for more tracking accuracy at the cost of compute time. We can also fuse these two concepts together to evaluate the need for more detection and tracking accuracy together, helping understand the upper-bound performance of real-world systems. A sample of the images available in the dataset is shown in Figure \ref{fig:sampleImages}. As can be seen from the figure, these are the kind of images one would expect from a fully automated system with detection and tracking modules working in a real-world surveillance environment. As noted in Table ~\ref{tab:dataset}, the Airport dataset suffers from all challenging attributes except low resolution. That is because relatively small detections are rejected by the person detector.

\subsection{Evaluation protocol}
\label{sec:bench:eval}
\subsubsection{Datasets, and training and testing splits.} Based on the number of images for each probe person, we categorize the datasets into either the single-shot or multi-shot setting. We employ the single-shot evaluation protocol for VIPeR, GRID, 3DPeS, DukeMTMC4ReID, CUHK01, CUHK02, CUHK03, HDA+, Market1501, and Airport. For the other 7 datasets, we employ the multi-shot evaluation protocol. In the Airport dataset, we fix one of the 6 cameras as the probe view and randomly pick paired people from 20 of the 40 short clips as the training set, with the rest forming the testing set. In the case of CUHK03, DukeMTMC4ReID, GRID, HDA+, and Market1501, we use the partition files provided by the respective authors. In particular, for the CUHK03 dataset, as noted in the previous section, we only use the ``detected" bounding boxes in all reported experiments. In RAiD, we fix camera 1 as the probe view, resulting in three sub-datasets, RAiD-12, RAiD-13, and RAiD-14, corresponding to the 3 possible gallery views. RAiD-12 has 43 people, of which we use 21 people to construct the training set and the rest to construct the testing set. The other two sub-datasets have 42 people each, which we split into equal-sized training and testing sets. In WARD, we fix camera 1 as the probe view, resulting in two sub-datasets, WARD-12 and WARD-13, corresponding to the 2 possible gallery views. Both these sub-datasets have 70 people each. We split VIPeR, CUHK01, CUHK02, GRID, CAVIAR, 3DPeS, PRID, WARD-12, WARD-13 and iLIDSVID into equal-sized training and testing sets. SAIVT-38 has 99 people, of which we use 31 people for training and the rest for testing. SAIVT-58 has 103 people, of which we use 33 people for training and the rest for testing. We note that in the cases of iLIDSVID, PRID, and SAIVT, the split protocol used here is the same as in previous works that propose multi-shot re-id algorithms \cite{wang2014person, wang2016personpami,liu2015spatio,mclaughlin2016recurrent,karanam2017person} to ensure evaluation consistency. Finally, for each dataset, we use 10 different randomly generated training and testing sets and report the overall average results.

\subsubsection{Evaluation framework.} In the single-shot evaluation scheme, for each dataset, we consider two aspects: type of feature and type of metric learning algorithm. We evaluate all possible combinations of the 11 different features and 18 different metric learning algorithms listed in Table \ref{tab:featMetric}. Since we also evaluate four different kernels for the kernelized algorithms, the total number of algorithm combinations is 276.\footnote{We evaluate only linear and exp kernels for LDFV, GOG, IDE-CaffeNet, IDE-ResNet, and IDE-VGGNet.}  In the multi-shot evaluation scheme, we consider three aspects: type of feature, type of metric learning algorithm, and type of ranking algorithm. Additionally, we consider two evaluation sub-schemes: using the average feature vector as the data representative (called AVER), and clustering the multiple feature vectors for each person and considering the resulting cluster centers as the representative feature vectors for each person (called CLUST). AVER effectively converts each dataset into an equivalent single-shot version. However, in the case of CLUST, we do not consider kernelized metric learning algorithms and other non-typical algorithms such as RankSVM and SVMML because only AVER can be employed to rank gallery candidates. Consequently, we use the remaining 9 metric learning algorithms and the baseline $l_{2}$ method, in which we use the features in the original space without any projection. These 10 algorithms are used in combination with the 11 different features and 4 different ranking algorithms. In total, we evaluate 646 different algorithm combinations for each multi-shot dataset. 

\subsubsection{Implementation and parameter details.} 
We normalize all images of a particular dataset to the same size, which is set to $128\times48$ for VIPeR, GRID, CAVIAR and 3DPeS and 128$\times$64 for all other datasets. To compute features, we divide each image into 6 horizontal rectangular strips. In the case of LOMO, since the patches are fixed to be square-shaped, we obtain 12 patches for a $128\times48$ image and 18 patches for a 128$\times$64 image. 

In the case of IDE-ResNet and IDE-VGGNet, we resize each image to 224$\times$224 pixels following \cite{simonyan2014very}. For IDE-CaffeNet, we resize each image to 227$\times$227. We start training with a model pre-trained on the ImageNet dataset and train the fully connected layers fc7 and fc8 from scratch. The number of output units in the fc7 layer is set to 4096 for IDE-VGGNet and IDE-CaffeNet, and 2048 for IDE-ResNet. Since we consider each person to be a different class, we set the number of output units in the fc8 layer to the number of unique people in our training set. Depending on the training split, this number varies from 2560 to 2580. Once the model is trained, we use the output of the fc7 layer as the image descriptor, giving a 4096-dimensional feature vector in the case of IDE-VGGNet and IDE-CaffeNet, and a 2048-dimensional feature vector in the case of IDE-ResNet.

In metric learning, we set the projected feature space dimension to 40. We set the ratio of the number of negative to positive pairwise constraints to 10.\footnote{This is set to 1 for kPCCA and rPCCA on Market1501 due to system memory issues.} In the case of CLUST, we set the number of clusters to 10, which we determine using the k-means algorithm. 

\section{Results and Discussion}
\label{sec:bench:results}

\begin{figure*}
\includegraphics[width=0.35\textwidth]{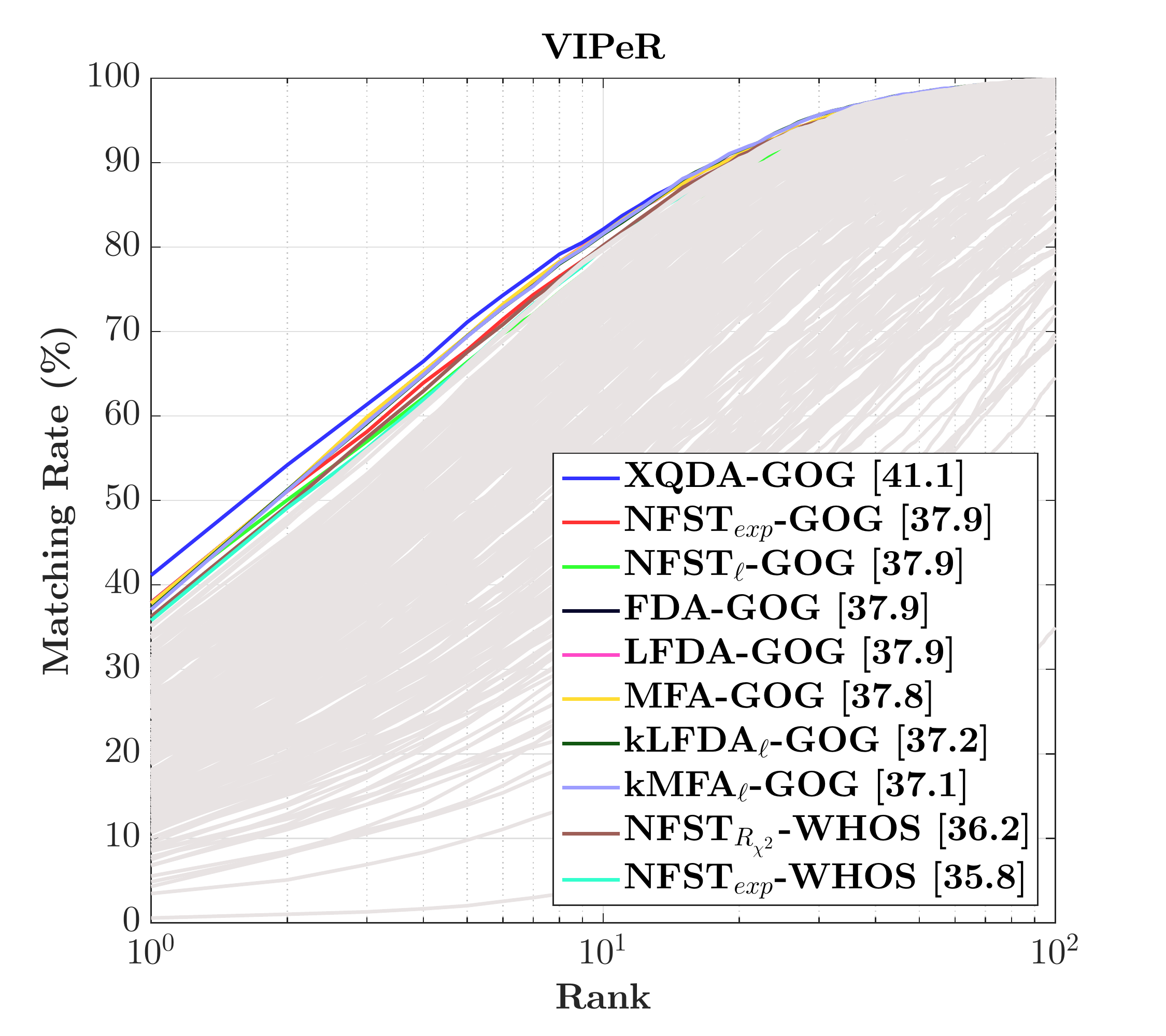}  
\includegraphics[width=0.35\textwidth]{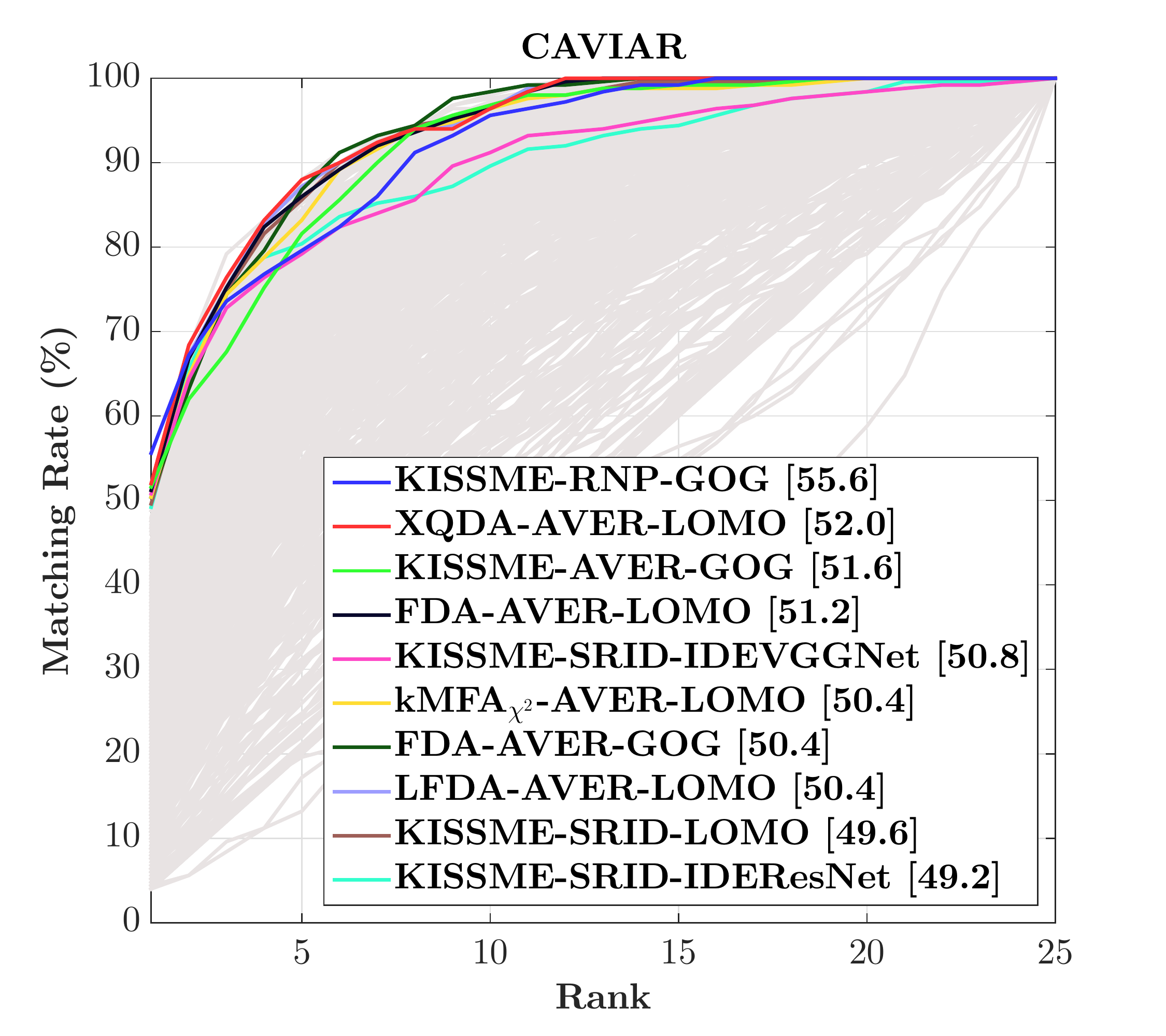} 
\caption{CMC curves for the single-shot dataset VIPeR and the multi-shot dataset CAVIAR. The algorithmic combinations with the ten best rank-1 performances (indicated in the legend) are shown in color and all the others are shown in gray. CMC curves for all other datasets can be found in the supplementary material.}
\label{fig:allDataCMC}
\vspace*{-1em}
\end{figure*}

\begin{table}[!htbp]
\renewcommand{\arraystretch}{1.2}
\centering
\caption{Top performing algorithmic combinations on each dataset, where we show the re-id performance (\%) at ranks 1, 5, and 10. Read as feature-metric for single-shot and feature-metric-ranking for multi-shot.}
\label{tab:bestAlgoEachDataset}
\scalebox{0.75}{
\begin{tabular}{|c|c|c|c|c|}
\hline
Datasets   & Best Combination        		& 1    & 5     & 10       \\ \hline \hline
VIPeR      & GOG-XQDA              		& 41.1 & 71.1  & 82.1   \\
\hline
GRID      & IDE-ResNet-KISSME           & 26.6 & 43.1  & 50.9    \\
\hline
3DPeS      & IDE-ResNet-NFST$_{exp}$    & 53.4 & 77.8  & 85.6    \\
\hline
CUHK01     & GOG-NFST$_{exp}$        	& 55.6 & 77.7  & 84.8    \\
\hline
CUHK02     & GOG-NFST$_{exp}$        	& 57.9 & 79.3  & 85.7    \\
\hline
CUHK03     & GOG-kLFDA$_{exp}$        	& 62.1 & 88.7  & 94.2    \\
\hline
HDA+       & IDE-ResNet-NFST$_{\ell}$   & 84.1 & 84.5  & 85.8    \\
\hline
Market1501 & IDE-ResNet-NFST$_{exp}$ 	& 64.3 & 80.9  & 86.2    \\
\hline
Airport    & IDE-ResNet-NFST$_{exp}$    & 42.7 & 67.5  & 76.0   \\ 
\hline
DukeMTMC4ReID    & IDE-ResNet-NFST$_{exp}$  & 54.6 & 68.6  & 73.7   \\ \hline \hline
PRID      & GOG-KISSME-SRID      			& 91.5 & 97.8  & 98.8   \\
\hline
V47    	& IDE-ResNet-KISSME-RNP        			& 100.0 & 100.0  & 100.0  \\
\hline
CAVIAR     & GOG-KISSME-RNP        			& 55.6 & 79.6  & 95.6   \\
\hline
WARD-12   & GOG-KISSME-SRID          			& 99.7 & 100.0 & 100.0  \\
\hline
WARD-13   & GOG-KISSME-ISR      			& 97.7 & 98.6  & 99.1  \\
\hline
SAIVT-38  & GOG-KISSME-SRID      			& 96.5 & 100.0  & 100.0  \\
\hline
SAIVT-58  & GOG-KISSME-RNP      			& 72.6 & 89.9  & 93.0    \\
\hline
RAiD-12   & IDE-ResNet-KISSME-AHISD      			& 100.0 & 100.0 & 100.0  \\
\hline
RAiD-13   & GOG-KISSME-SRID         			& 81.9 & 94.8  & 96.2  \\
\hline
RAiD-14   & GOG-KISSME-SRID      			& 95.7 & 96.2  & 99.5  \\
\hline
iLIDSVID    & GOG-KISSME-SRID         			& 75.7 & 90.1  & 93.6   \\
\hline

\end{tabular}
}
\vspace{-1em}
\end{table}

We first summarize the results of the overall evaluation, and then discuss several aspects of these results in detail. 

The overall cumulative match characteristic (CMC) curves for two representative single- and multi-shot datasets are shown in  in Figure~\ref{fig:allDataCMC}. The CMC curve is a plot of the re-identification rate at rank-k. The individual performance of each algorithm combination on all datasets as well as complete CMC curves can be found in the supplementary material. As can be seen from the CMC curves, the ``spread'' in the performance of the algorithms for each dataset is huge, indicating the progress made by the re-id community over the past decade. However, on most datasets, the performance is still far from the point where we would consider re-id to be a solved problem.  
In Table~\ref{tab:bestAlgoEachDataset}, we summarize the overall CMC curves by reporting the algorithm combination that achieved the best performance on each dataset as measured by the rank-1 performance. We note that IDE-ResNet \cite{he2015deep} and GOG \cite{matsukawa2016hierarchical} perform the best among the 11 evaluated feature extraction algorithms, with them being a part of the best performing algorithm combination in 6 of the 10 single-shot and all the 11 multi-shot datasets respectively. 
In the case of multi-shot evaluation, the combination of KISSME \cite{koestinger2012large} as the metric learning algorithm and SRID \cite{karanam2015sparse} as the multi-shot ranking algorithm is the best performing algorithm combination, with it resulting in the best performance on 6 of the 11 datasets.

In general, we observe that the algorithms give better performance on multi-shot datasets than on single-shot datasets. While this may be attributed to multi-shot datasets having more information in terms of the number of images per person, it is important to note that the single-shot datasets considered here generally have a significantly higher number of people in the gallery. It is quite natural to expect re-id performance to go down as the number of gallery people increases because we are now searching for the person of interest in a much larger candidate set.

\subsection{Single shot analysis: features and metric learning}
\label{sec:bench:results:single}
\begin{figure*}[!htbp]
\begin{center}
\begin{tabular}{c c c}
\includegraphics[width=0.33\textwidth]{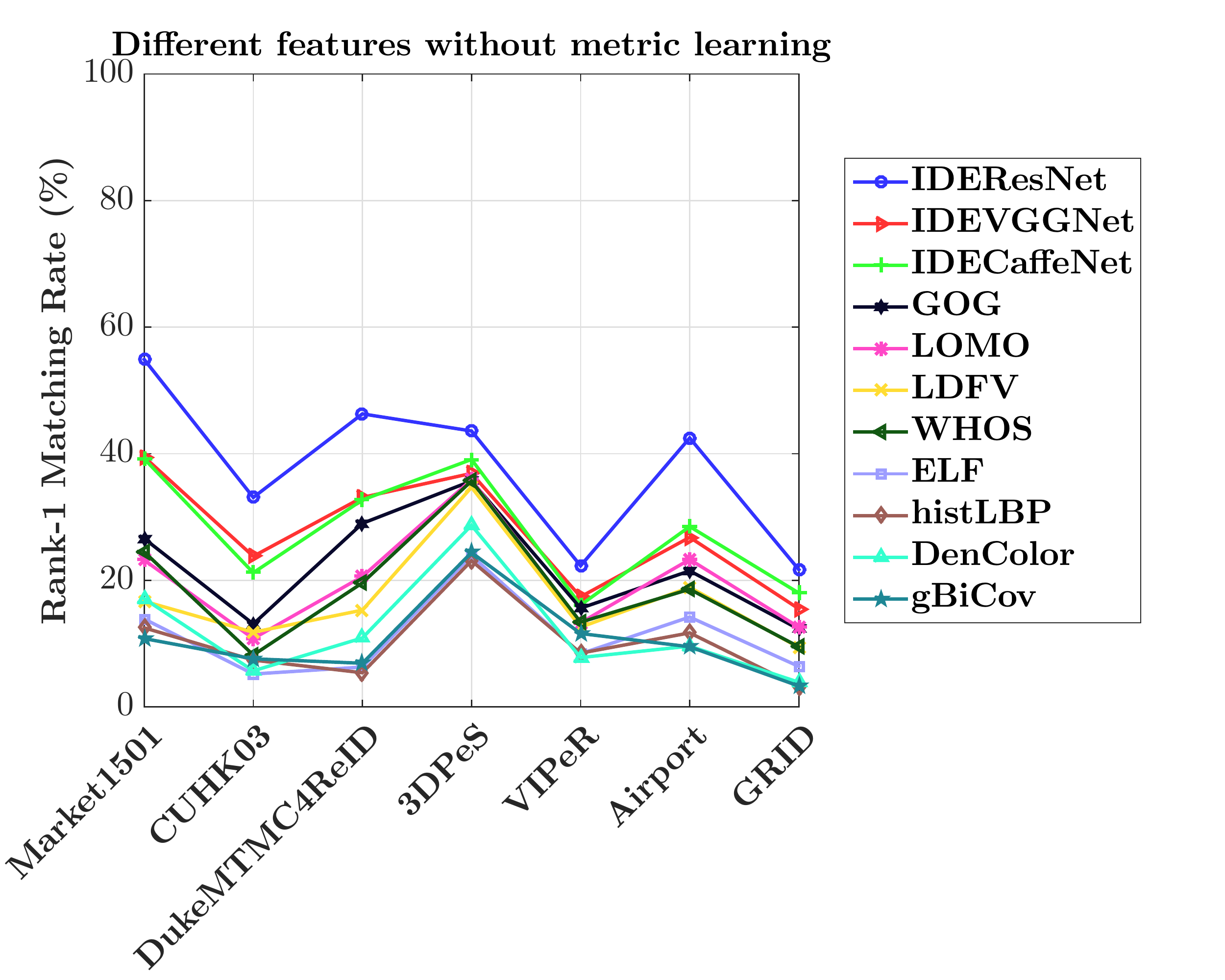} &
\includegraphics[width=0.33\textwidth]{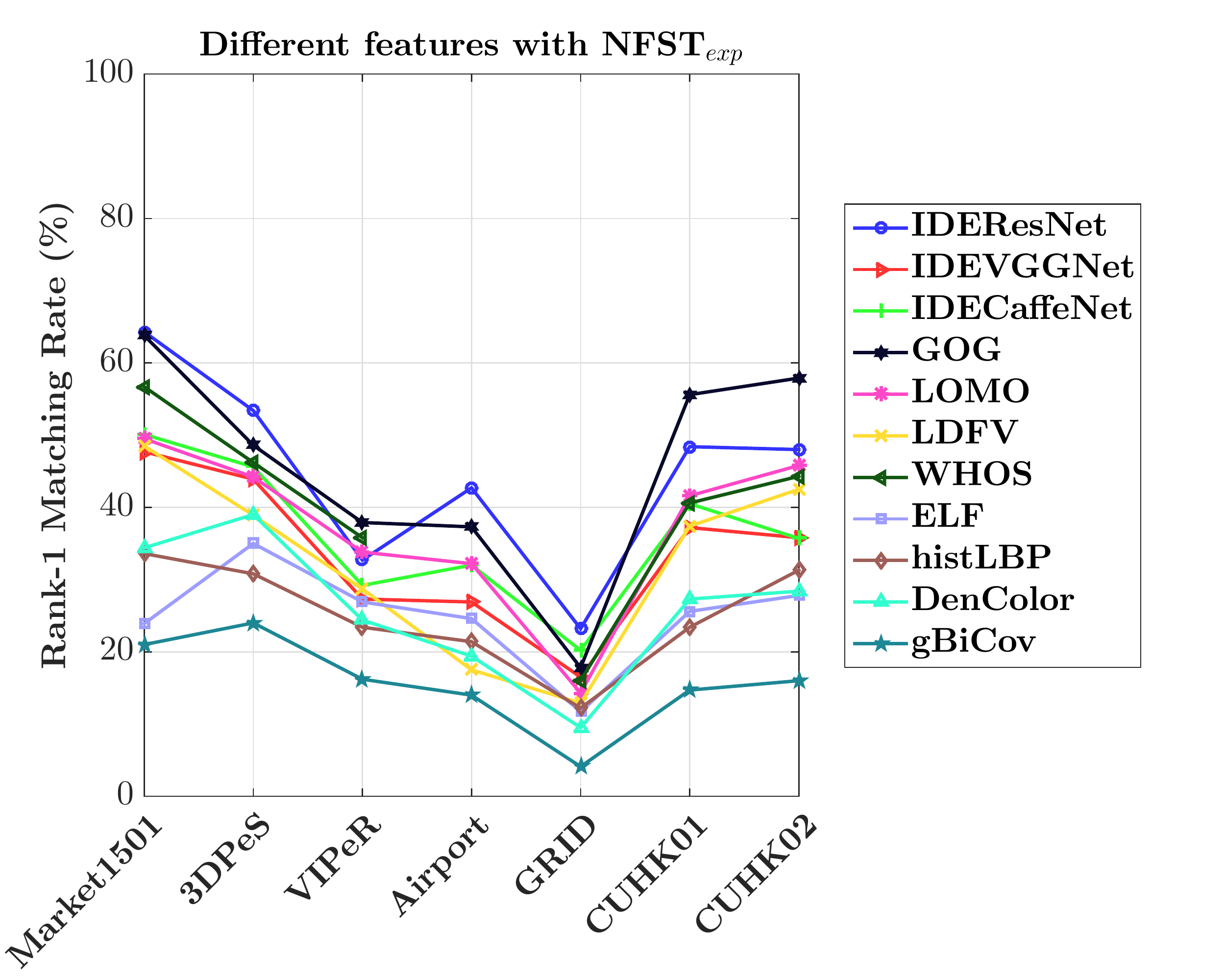} &
\includegraphics[width=0.33\textwidth]{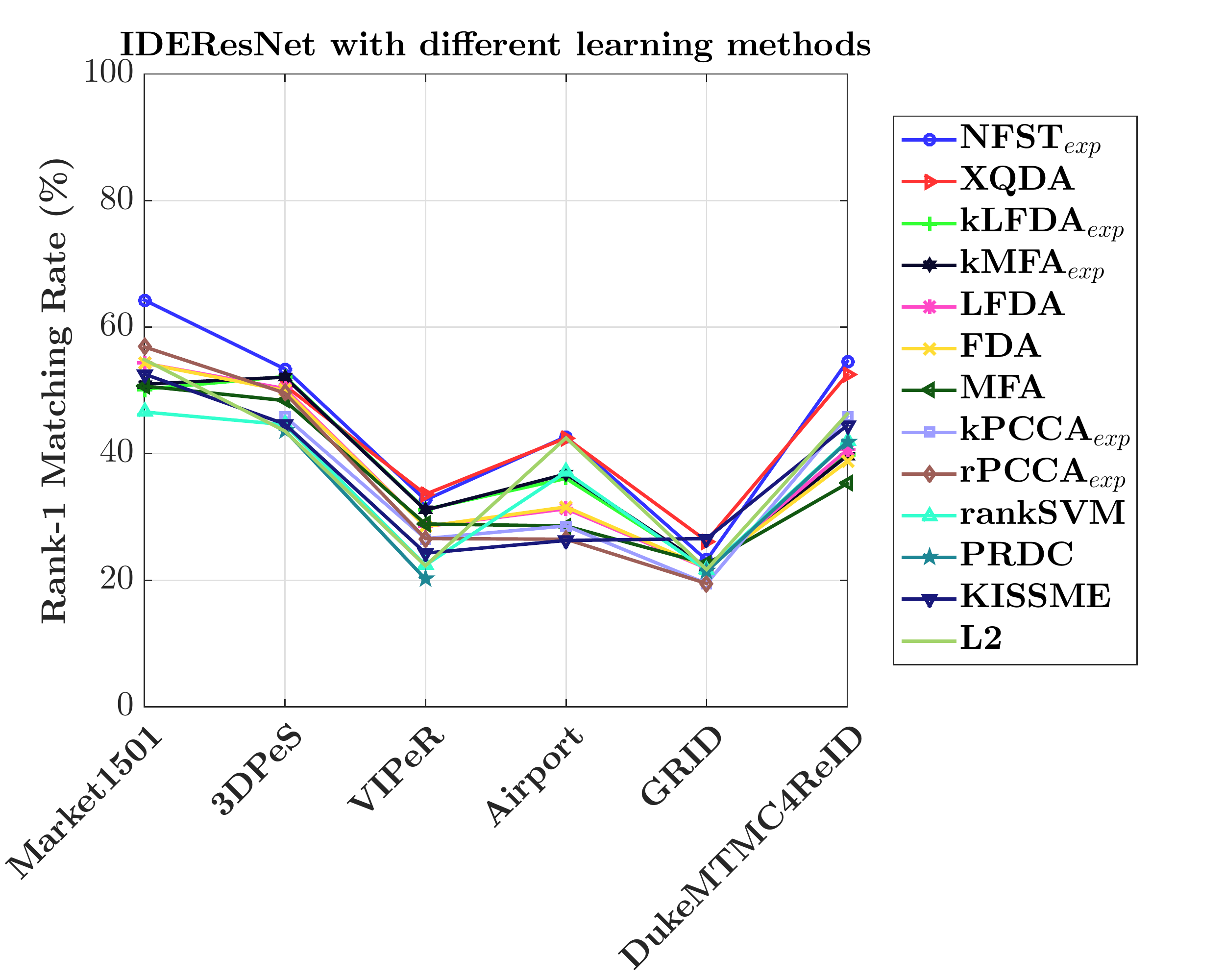} \\
(a)&(b)&(c)
\end{tabular}
\caption{Rank-1 results for single shot datasets illustrating the impact of IDE-ResNet and NFST$_\text{exp}$.}
\label{fig:single_feat_algo}
\end{center}
\vspace*{-1.5em}
\end{figure*}

Single-shot re-id involves two key aspects: features and metric learning. In this section, we isolate the impact of the best performing algorithm in these two areas. First, we note that IDE-ResNet is the best performing feature extraction algorithm in our evaluation. To corroborate this observation, we study the impact of IDE-ResNet in comparison with other feature extraction algorithms both in the presence as well as the absence of any metric learning. In the first experiment, we use the baseline Euclidean distance to rank gallery candidates in the originally computed feature space, which can be regarded as an unsupervised method. As can be noted from the results shown in Figure \ref{fig:single_feat_algo}(a)\footnote{In the graph, we only show results on 7 datasets for brevity. Please consult supplementary material for complete results.}, IDE-ResNet gives the best performance on all the  10 datasets. 

Next, we study how IDE-ResNet performs in comparison with other features in the presence of metric learning. In this experiment, we fix NFST$_\text{exp}$ as our metric learning algorithm and rank gallery candidates using all the 11 evaluated feature extraction algorithms. The rank-1\footnote{Complete CMC curves can be found in the supplementary material.} results for this experiment are shown in Figure~\ref{fig:single_feat_algo}(b). As can be noted from the graph, IDE-ResNet gives the best performance on 4 of the 7 datasets shown in the figure, with GOG giving the best performance on the remaining 3 datasets. 

These experiments show that IDE-ResNet is indeed the best performing feature extraction algorithm. This should not be surprising given the powerful modeling and generalization ability of the ResNet architecture, which is also evidenced by its strong performance in other computer vision domains and applications \cite{he2015deep}.

Here, we also note that GOG, despite being a hand-crafted feature extraction algorithm, results in competitive respectable performance. This is because color and texture are the most descriptive aspects of a person image and GOG describes the global color and texture distributions using a local Gaussian distributions of pixel-level features. Another critical reason for the success of GOG is the hierarchical modeling of local color and texture structures. This is a critical step because typically a person's clothes consists of local parts, each of which has certain local properties. 

Next, we analyze the performance of different metric learning algorithms\footnote{A discussion on the training time of these algorithms is provided in the supplementary material.}, in the context of IDE-ResNet, the best performing feature extraction algorithm. In this experiment, we fix IDE-ResNet as the feature extraction algorithm and study how different metric learning algorithms perform. The results of this experiment are shown in Figure~\ref{fig:single_feat_algo}(c), from which we can note that NFST$_\text{exp}$ gives the best performance on Market1501, DukeMTMC, 3DPeS, and Airport, with XQDA and kLFDA not being too far behind. These results further corroborate what we observe in Table~\ref{tab:bestAlgoEachDataset}, with NFST, kLFDA, and XQDA being among the best performing metric learning algorithms.

From the above discussion, we can infer the following: while NFST$_\text{exp}$ gives the best overall performance, kLFDA and XQDA also emerge as strong and competitive metric learning algorithms. It is interesting to note that all these three algorithms learn the distance metric by solving some form of generalized eigenvalue decomposition problems, similar to traditional Fisher discriminant analysis. While kLFDA and XQDA directly employ Fisher-type objective functions, NFST uses the Foley-Shannon transform \cite{foley1975optimal}, which is very closely related to the Fisher discriminant analysis. This suggests that the approach of formulating discriminant objective functions in terms of data scatter matrices is most suitable to the re-id problem.

\subsection{Multi-shot analysis: features, metric learning, and ranking}
\label{sec:bench:results:multi}

\begin{figure*}
\vspace*{-2em}
\centering
\vspace*{-1em}
\begin{tabular}{c c}
\includegraphics[width=0.4\textwidth]{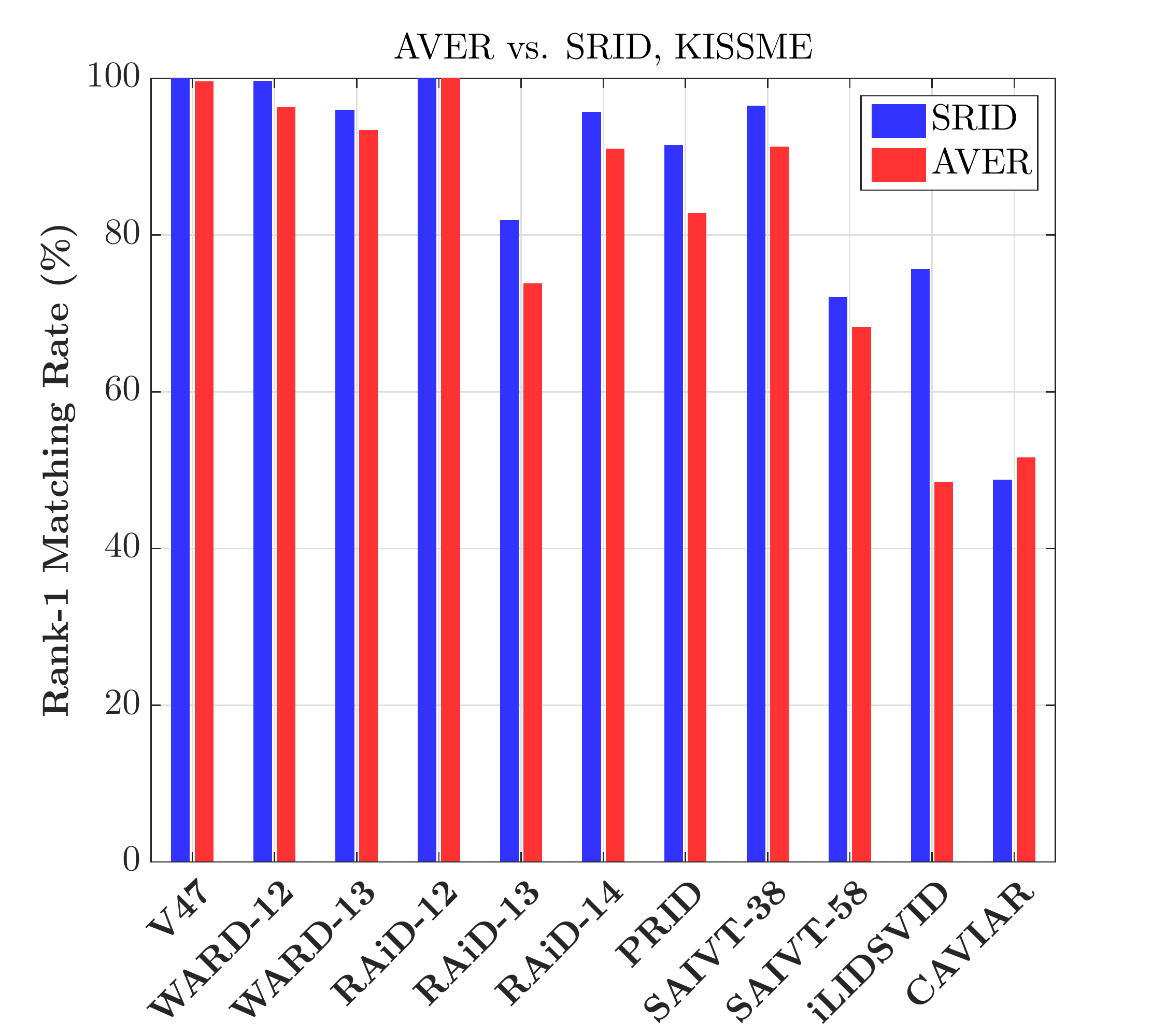}  &
\includegraphics[width=0.4\textwidth]{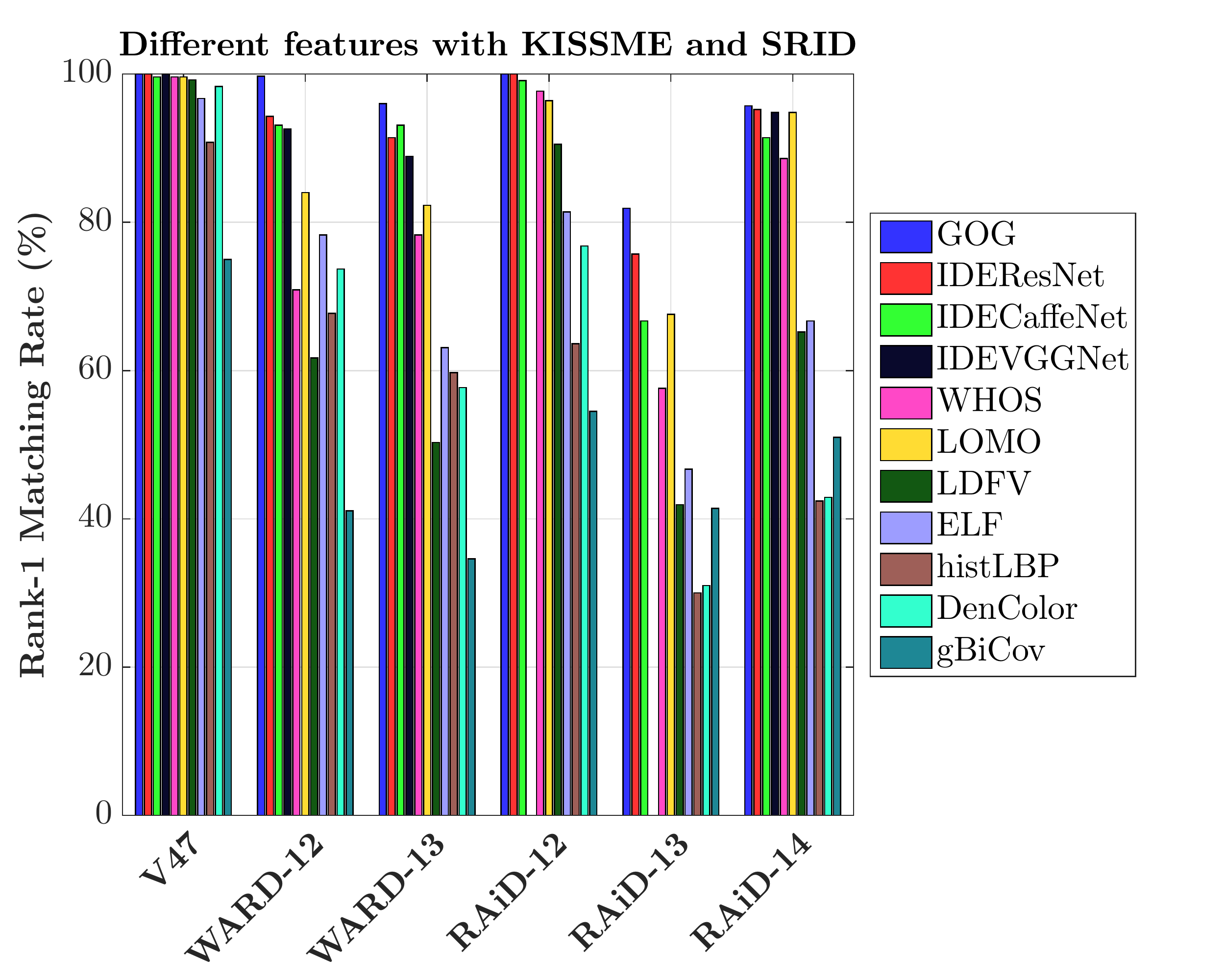}  \\
\includegraphics[width=0.4\textwidth]{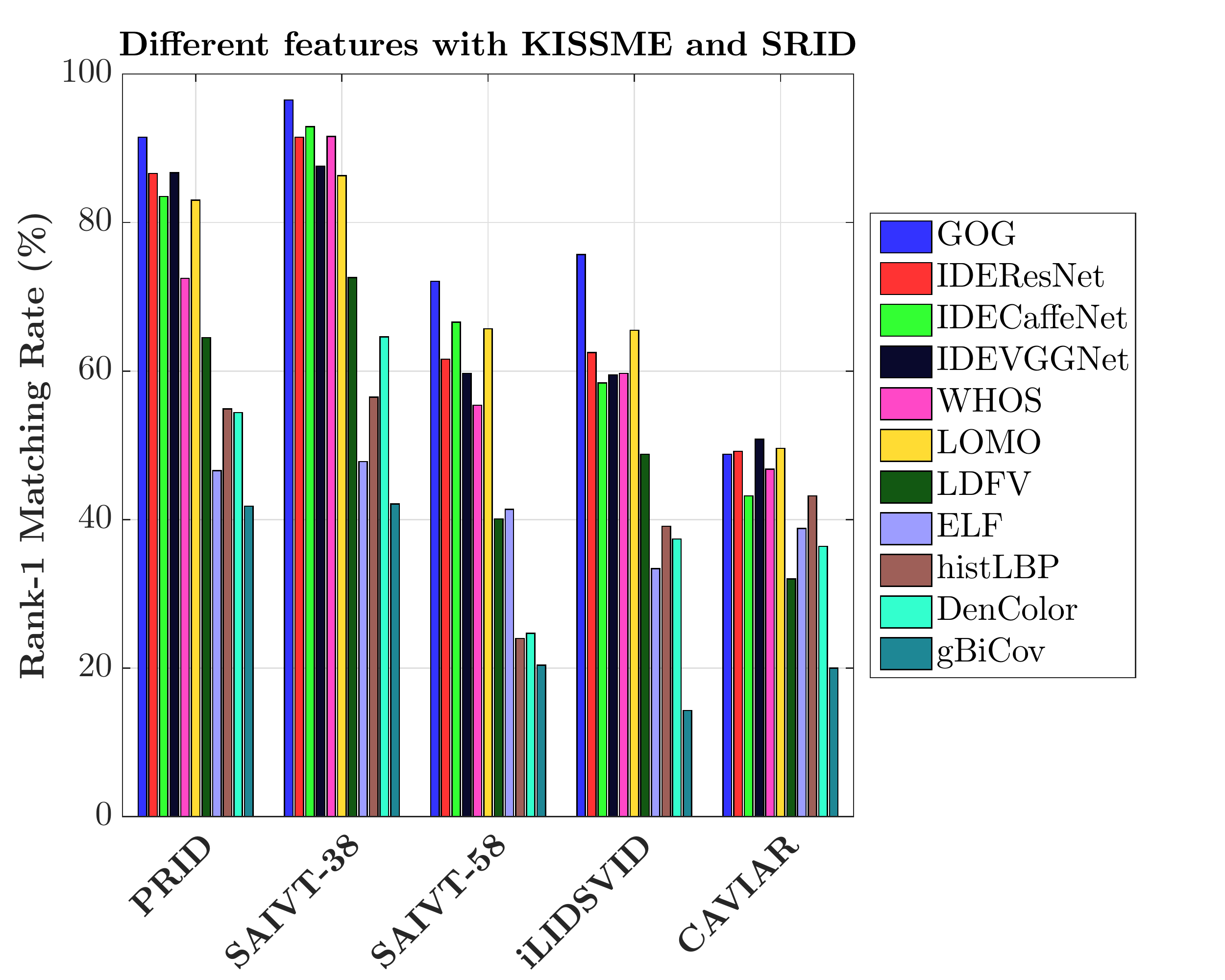} &
\includegraphics[width=0.4\textwidth]{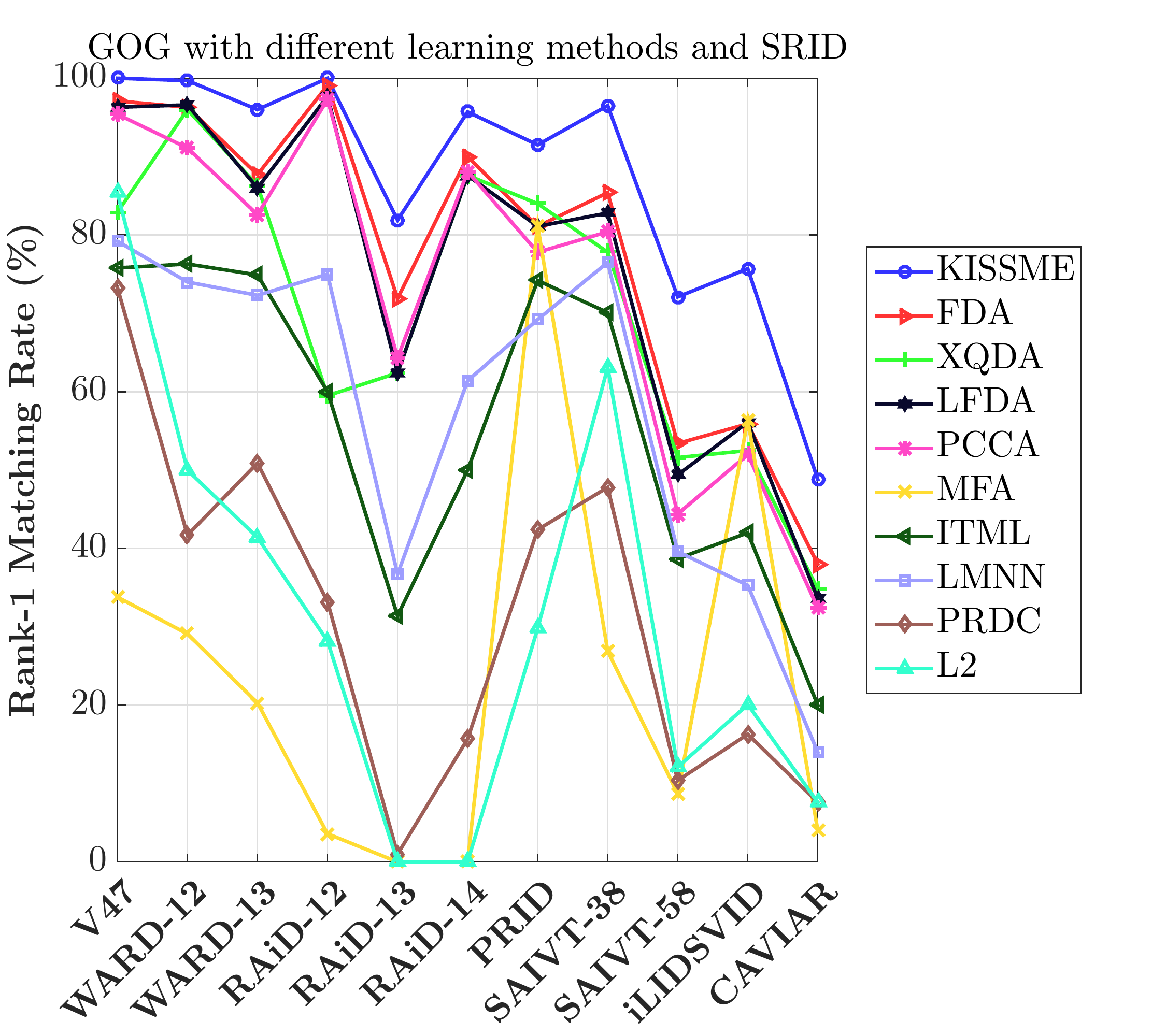} \\
(a)  & (b) \\ (c) & (d)
\end{tabular}
\caption{(a): Rank-1 performance on multi-shot datasets, illustrating the impact of the best performing multi-shot ranking algorithm, SRID over AVER, naive feature averaging. (b)-(d) Rank-1 performance on multi-shot datasets comparing various feature extraction and metric learning algorithms with SRID as the ranking algorithm.}
\label{fig:multi_shot_srid_aver}
\vspace*{-1.2em}
\end{figure*}

Multi-shot re-id involves three aspects: features, metric learning, and ranking. As noted previously, GOG, KISSME, and SRID emerged as the best performing algorithmic combination. On all the datasets, as expected, a custom ranking algorithm resulted in the best performance, with SRID performing the best on 6 of these 11 datasets. In this section, we provide further empirical results analyzing the impact of using a multi-shot ranking algorithm. To this end, we fix GOG as the feature extraction scheme. 

First, we evaluate the impact of using a multi-shot ranking algorithm instead of AVER. Here, we compare the performance of GOG-KISSME-AVER and GOG-KISSME-SRID. The results are shown in Figure~\ref{fig:multi_shot_srid_aver}(a). As can be noted from the graph, with the exception of CAVIAR, SRID generally gives superior performance when compared to AVER. This, and our observations from Table~\ref{tab:bestAlgoEachDataset}, suggest that using a multi-shot ranking algorithm that exploits the inherent structure of the data instead of naive feature averaging will give better performance. Furthermore, we also note that a multi-shot ranking algorithm in itself is not sufficient to give good performance because that would be purely an unsupervised approach. Combining a metric learning algorithm with the ranking technique adds a layer of supervision to the overall framework and will provide a significant performance boost. 

Next, we analyze the performance of the feature extraction and metric learning algorithms and compare the observed trends with those in the single-shot case. In the feature extraction case, we fix SRID as the ranking algorithm and KISSME as the metric learning algorithm. The rank-1 results for this experiment are shown in Figure \ref{fig:multi_shot_srid_aver}(b)-(c). We see very clear trends in this case, with GOG giving the best results on all the 11 datasets. These results are not surprising given the strong performance of GOG in the single-shot case. In the metric learning case, we fix SRID as the ranking algorithm and GOG as the feature extraction algorithm, with Figure \ref{fig:multi_shot_srid_aver}(d) showing the rank-1 results. We see very clear trends in this case as well, with KISSME giving the best results across all datasets.

\subsection{Additional observations}
\label{sec:bench:results:misc}
\begin{figure*}
\vspace*{-5em}
\centering
\begin{tabular}{c c}
\vspace*{-0.1em}
\includegraphics[width=0.4\textwidth]{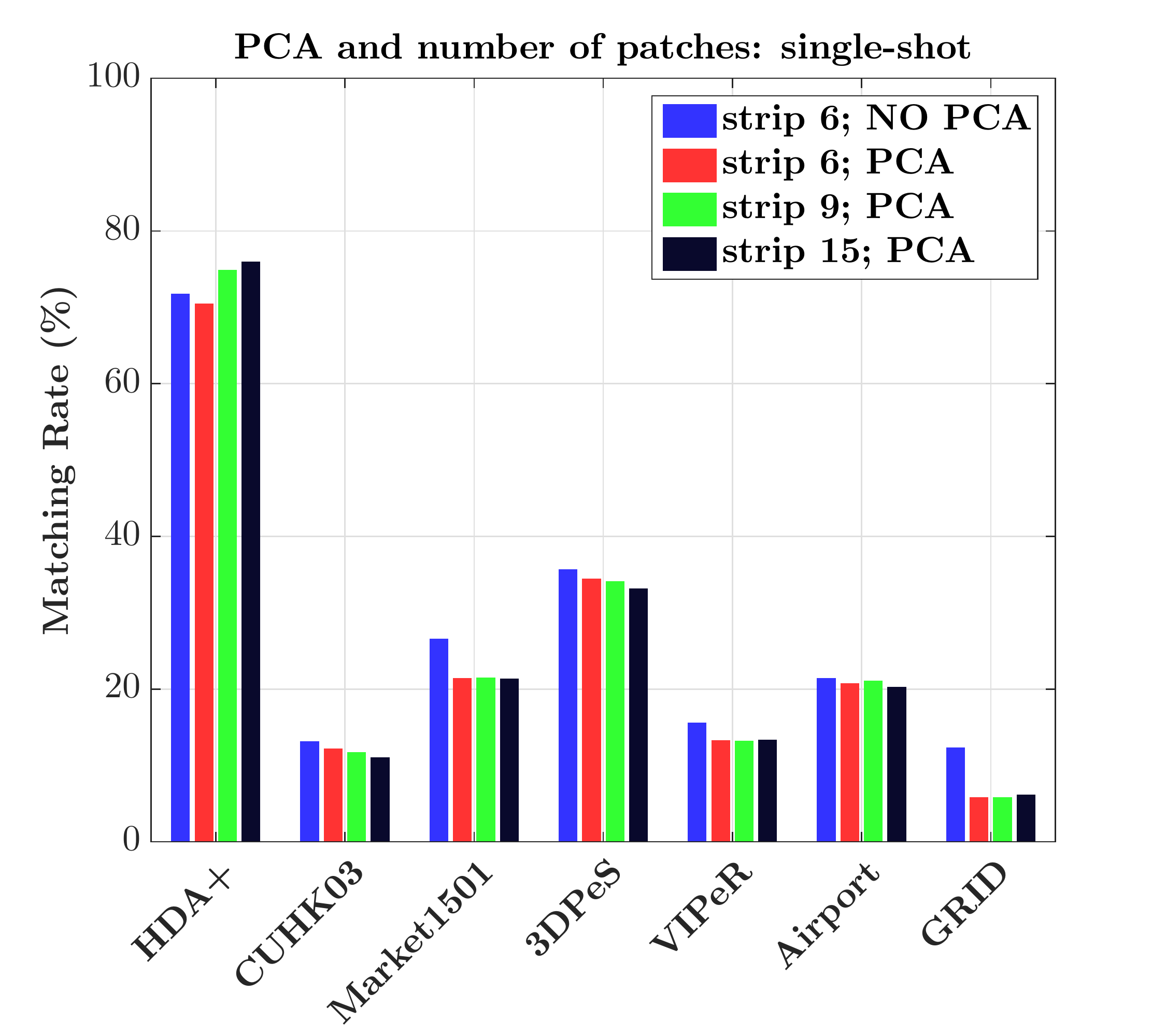}  &
\includegraphics[width=0.4\textwidth]{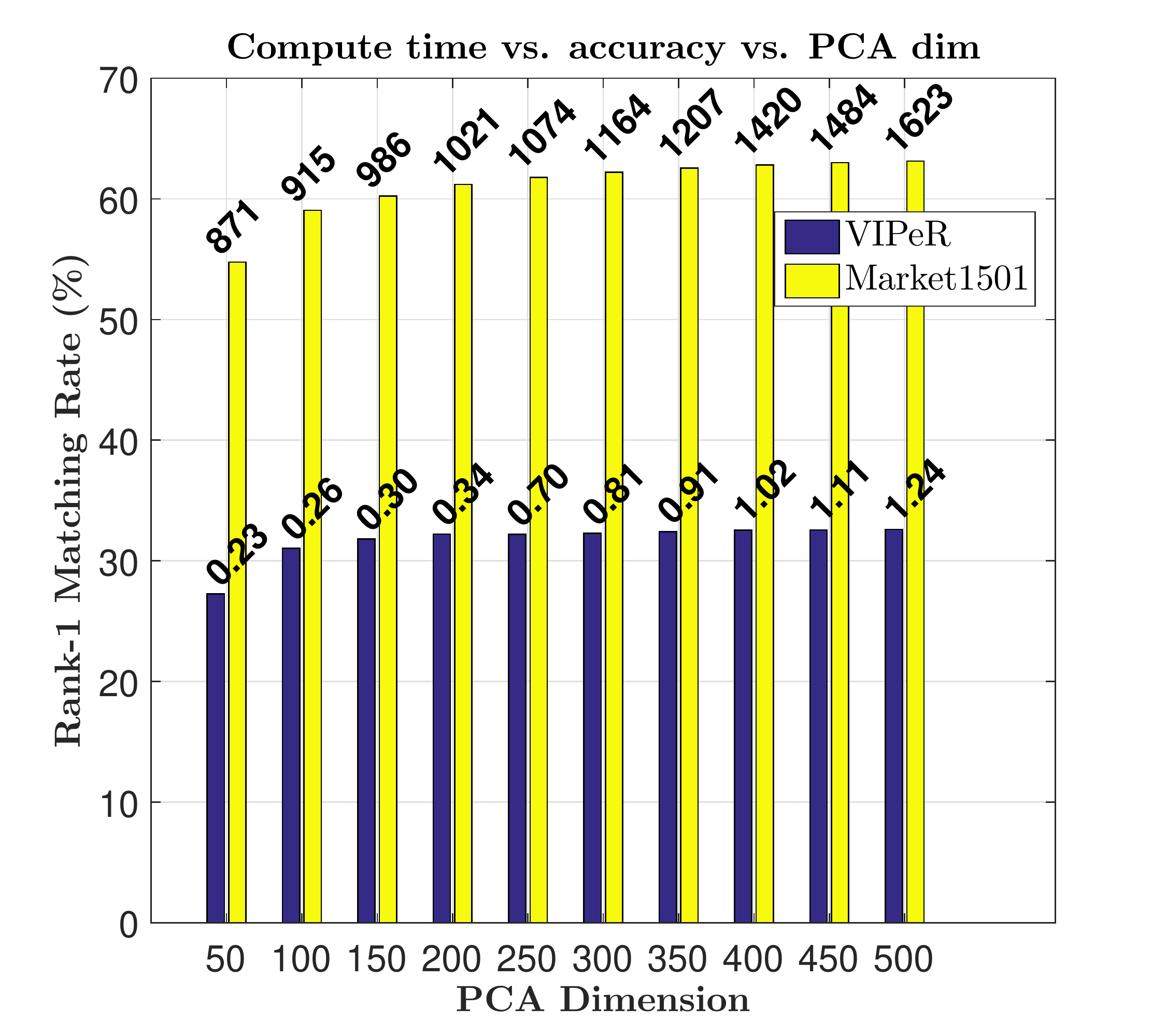} \\ 
(a) & (b) 
\end{tabular}
\caption{Rank-1 performance on (a) single-shot datasets illustrating the impact of PCA and number of strips. (b) Rank-1 and training time (in seconds) for VIPeR and Market1501 for various values of the PCA dimension.} 
\label{fig:misc_expt_single_multi}
\vspace*{-1.2em}
\end{figure*}

\begin{table*}[]
\centering
\caption{Mean rank-1 performance across all single- and multi-shot datasets with respect to various attributes and features. The best deep learning feature is shown in red and the best and second best hand-crafted features are shown in green and blue respectively.}
\label{tab:attr}
\begin{adjustbox}{max width=\linewidth}
\begin{tabular}{c|c|ccc||cccccccc}
\hline
                              & Attr & IDERes                          & IDEVGG & IDECaffe & GOG                                  & LOMO                                 & LDFV                                 & WHOS                                 & ELF  & HistLBP & DenColor & gBiCov \\ \hline
                              & VV   & {\color[HTML]{FE0000} \textbf{40.7}} & 30.4        & 29.2          & {\color[HTML]{008000} \textbf{26.6}} & 18.3                                 & {\color[HTML]{3531FF} \textbf{19.6}} & 17.2                                 & 11.7 & 9.0     & 11.7     & 15.5   \\
                              & IV   & {\color[HTML]{FE0000} \textbf{45.8}} & 36.3        & 33.5          & {\color[HTML]{008000} \textbf{34.7}} & 22.6                                 & {\color[HTML]{3531FF} \textbf{26.7}} & 20.7                                 & 16.6 & 12.0    & 15.8     & 23.1   \\
                              & BC   & {\color[HTML]{FE0000} \textbf{36.8}} & 25.1        & 26.4          & {\color[HTML]{008000} \textbf{20.9}} & {\color[HTML]{3531FF} \textbf{18.8}} & 14.5                                 & 15.9                                 & 9.0  & 6.8     & 8.1      & 6.6    \\
                              & OCC  & {\color[HTML]{FE0000} \textbf{35.3}} & 23.8        & 24.4          & {\color[HTML]{008000} \textbf{19.3}} & {\color[HTML]{3531FF} \textbf{15.2}} & 13.3                                 & 13.7                                 & 6.7  & 5.8     & 6.9      & 7.5    \\
\multirow{-5}{*}{Single-shot} & RES  & {\color[HTML]{FE0000} \textbf{38.3}} & 27.5        & 28.6          & {\color[HTML]{008000} \textbf{19.5}} & {\color[HTML]{3531FF} \textbf{17.9}} & 13.1                                 & 17.0                                 & 10.2 & 7.9     & 10.5     & 7.1    \\ \hline
                              & VV   & {\color[HTML]{FE0000} \textbf{49.1}} & 46.7        & 37.9          & {\color[HTML]{008000} \textbf{36.7}} & {\color[HTML]{3531FF} \textbf{20.2}} & 11.9                                 & 15.7                                 & 10.5 & 8.2     & 9.2      & 8.4    \\
                              & IV   & {\color[HTML]{FE0000} \textbf{65.1}} & 61.5        & 50.7          & {\color[HTML]{008000} \textbf{51.5}} & {\color[HTML]{3531FF} \textbf{24.1}} & 13.7                                 & 21.7                                 & 11.2 & 9.1     & 7.3      & 8.6    \\
                              & BC   & {\color[HTML]{FE0000} \textbf{61.0}} & 52.3        & 53.6          & {\color[HTML]{008000} \textbf{43.7}} & 17.2                                 & 14.5                                 & {\color[HTML]{3531FF} \textbf{34.2}} & 5.2  & 6.6     & 9.2      & 12.4   \\
                              & OCC  & {\color[HTML]{FE0000} \textbf{42.0}} & 37.4        & 27.5          & {\color[HTML]{008000} \textbf{23.2}} & 6.8                                  & 5.5                                  & {\color[HTML]{3531FF} \textbf{12.9}} & 5.7  & 3.1     & 8.3      & 5.1    \\
\multirow{-5}{*}{Multi-shot}  & RES  & {\color[HTML]{FE0000} \textbf{34.0}} & 26.8        & 24.4          & {\color[HTML]{008000} \textbf{28.4}} & {\color[HTML]{3531FF} \textbf{25.2}} & 12.4                                 & 14.0                                 & 18.0 & 15.6    & 13.6     & 11.2 \\ \hline
\end{tabular}
\end{adjustbox}
\vspace*{-1.2em}
\end{table*}

In this section, we report additional empirical observations. Most contemporary feature extraction schemes produce high-dimensional data, introducing significant computational overhead. To this end, we analyze the impact of an unsupervised dimensionality reduction scheme, principal component analysis (PCA). We fix GOG, features with the highest dimensionality in our evaluation framework, as the feature extraction scheme and perform experiments with and without PCA. We set the dimension of the PCA-reduced space to 100. The results are shown in the first two bars (pink and yellow) in Figure \ref{fig:misc_expt_single_multi}(a). The results without PCA are better than those with PCA on all the datasets shown in the graphs. This observation is not surprising given that PCA can result in the undesirable removal of the most discriminative features. 

All hand-crafted feature extraction algorithms use some form of localized feature computation by dividing the image into pre-defined strips. Here, we analyze the impact of the number-of-strips parameter. To this end, we perform experiments with 6, 9, and 15 horizontal strips in the best hand-crafted feature extraction algorithm, GOG, with Euclidean distance as the metric in the single-shot case and Euclidean distance as the metric and AVER as the ranking strategy in the multi-shot case. The rank-1 results are shown in bars 2--5 in Figure \ref{fig:misc_expt_single_multi}(a) \footnote{The rank-1 results for the multi-shot case are provided in the supplementary material.}. While it is reasonable to expect superior performance as we increase the number of strips, thereby increasing the feature space dimensionality, it is important to note that in this process, we may have fewer samples to estimate the Gaussians in each strip and also increase the amount of background/noise/non-informative content in the feature descriptor. We also note that there does not seem to be any significant performance variations as we increase the number of strips. Given the computational complexity involved in working with higher dimensional feature spaces due to increased number of strips, these results suggest that 6 strips, which is in fact the widely used number in the re-id community, seems to be a reasonable choice, giving better or close performance to the other choices in most cases. 

Finally, we also empirically study how re-id accuracy varies vis-a-vis computational requirements for various values of the PCA dimension. In Figure~\ref{fig:misc_expt_single_multi}(b), we show results of this experiment for values of PCA dimension ranging from 50 to 500 for a small-scale dataset, VIPeR, and a large-scale dataset, Market1501. The numbers on top of the bars are the training times in seconds. As can be noted from the results, as we increase the PCA dimension, the training time increases (quite substantially for the large-scale dataset), while the accuracy saturates beyond a certain value of the PCA dimension. This empirically substantiates the sufficiency of a relatively small value for performing dimensionality reduction using algorithms such as the PCA.

\subsection{Attribute-based analysis}
In this section, we analyze the performance of the different feature extraction schemes with respect to the different attributes used to characterize datasets in Table \ref{tab:dataset}. The goal of this experiment is to study which features are good in certain scenarios. To this end, we use Euclidean distance as the metric, and in the multi-shot case, AVER as the ranking algorithm, and report the mean rank-1 performance on all datasets for each attribute group. The results obtained are shown in Table \ref{tab:attr}. We observe the following trends from the results. In all the scenarios, IDE-ResNet resulted in the best performance, with IDE-VGGNet and IDE-CaffeNet following closely behind. For a few attributes, GOG gave competitive, albeit lower, performance when compared to the IDE features (e.g., VV, IV, and RES). 

While the IDE results are not surprising given the relatively strong supervision from annotated training data, hand-crafted features can provide us with insights on learning more re-id specific domain knowledge. While we only discuss the best performing hand-crafted algorithms- GOG, LDFV, and LOMO- these insights can be quite useful across the spectrum as researchers think about designing feature learning architectures. First, as opposed to other hand-crafted algorithms, these three methods explicitly model local pixel distributions. While this is intuitively obvious, we note that is an extremely important step in describing person images. Additionally, since viewpoint invariance is an extremely important attribute for any re-id descriptor, these results suggest that incorporating local region information and horizontal strip pooling, as done explicitly in both GOG and LOMO, is critical to achieve viewpoint invariant descriptors. Furthermore, we note that WHOS results in strong performance on BC and RES in the single-shot case, primarily due to the use of a mask that filters out background clutters, resulting in a more localized features representation, similar in spirit to the approach noted above.

\subsection{Impact of datasets on re-id research}
\label{sec:bench:results:misc:datasets}
Datasets play an extremely important role in validating a newly proposed algorithm. From Table~\ref{tab:bestAlgoEachDataset}, we note that the V47 dataset can be considered to be solved, with the best performing algorithm achieving a rank-1 performance of 100\%. However, the performance on other datasets is still far from ideal. These datasets therefore present opportunities and challenges to develop better algorithms. For instance, the performance on VIPeR is still very low despite it being the most popular dataset in the re-id community. The performance on GRID is the lowest (26.6\% at rank-1) and this is in part due to the presence of a large number of distractor people in the gallery. The newly proposed Airport dataset has the next lowest performance (42.7\% at rank-1). This is due to the presence of a large distractor set as well as the significant real-world challenges described in Section \ref{sec:newDataset}. These observations suggest that as the number of distractor people in the gallery increases, the performance of a re-id algorithm goes down. This is not surprising since now we have a larger set of people to compare the probe person against, leading to more avenues for the re-id algorithm to fail.  

Generally speaking, there are two key aspects that need to be considered while constructing new datasets: they have to be both  \textbf{large} and \textbf{realistic}. As noted above, the presence of a distractor set helps mimic the real-world nature to a certain extent. An important point we would like to emphasize relates to the notion of distractors as commonly used in constructing datasets; in most cases, these correspond to false detections provided by a person detector. While this is somewhat reasonable, in the real world, we have both false alarms and actual person images, so having a number of other unpaired person images, in addition to false positives from a person detector, is crucial. Given that end-users, mostly security personnel, will be searching for a person of interest among hundreds of thousands of people, having a re-id dataset this large would help to quickly scale-up re-id algorithms to become practically relevant. Furthermore, and crucially, datasets must be relevant to real-world application scenarios. Most current dataset construction mechanisms focus on capturing images of people under different camera views, illumination conditions, or other factors such as occlusion. While these factors are important, what is missing is more fundamental: since re-id primarily finds applications in crime detection and prevention, the perpetrator or the person of interest may re-appear in a different (e.g., changed clothes, hair style, etc.) appearance. Again, for re-id algorithms to be used in the real world, we need datasets that capture this subtle yet extremely important aspect. A more extensive discussion is provided in Section~\ref{sec:futureresearch}.

Our categorization of datasets according to their attributes also provides a useful reference point for constructing new datasets. We note that none of the 14 datasets have all 6 attributes. While MARS \cite{zheng2016mars}, a recently proposed dataset, has a large number of images, constructing datasets that are of the size of ImageNet, in terms of the number of people, positive examples, and  under an eclectic mix of conditions as noted above, will assist in the application of some of the recent algorithmic advances in feature learning using CNNs \cite{szegedy2015going,he2015deep}. We believe focusing on the aspects discussed in this section, and several others discussed in Section~\ref{sec:futureresearch}, when constructing new datasets, would help rapidly accelerate progress in person re-id.

\section{Insights and Recommendations for Researchers}
\label{sec:futureresearch}
Our systematic study of re-id algorithms across many datasets helps us characterize what re-id algorithms are currently capable of doing, as well as what is missing and can be done in the future. To this end, we here discuss insights gained from this exercise, as well as posit research directions and recommendations for re-id researchers that would help develop better algorithms.

We noted that IDE-ResNet resulted in the best performance among all the evaluated feature extraction methdods, with IDE-VGGNet and IDE-CaffeNet close behind. Again, this is expected due to powerful, generalizable features that CNNs are capable of learning giving enough supervision. While adapting more recent improvements in architecture design and feature learning \cite{huang2016densely,chen2017beyond} will naturally give better performance, we would, in particular, like to note the strong performance given by GOG. A primary reason for its success is the hierarchical modeling of local pixel distributions, first at the patch level and then at the strip level which is inspired by LOMO. There are two key takeaways from this observation: local features and hierarchical modeling. Intuitively, this should not be surprising since using local features helps mitigate potential issues with background clutter and noise, two challenges that are critical for re-id. Following recent advances in feature learning, a particularly promising research direction would be to learn \textbf{local patch representations}, which can then be aggregated into an image-level descriptor using existing aggregation schemes \cite{jegou2010aggregating}. The region proposal network of Faster R-CNN \cite{ren2015faster} is a potential candidate to generate local patch proposals and learn representations, which can be trained in an end-to-end fashion. There have been some recent efforts to this end  \cite{li2017learning,zhao2017spindle}, where specific architectures are designed to learn local body-region features. Strip-level pooling is another important aspect that is unique to the re-id problem. Because people are roughly vertically aligned in person images, translation-invariant pooling for each local strip (typically constructed horizontally) can help mitigate issues caused by viewpoint variations across cameras. Consequently, recent advances in learning translation invariant local features such as bilinear CNNs \cite{lin2015bilinear} or gated CNNs with specially designed convolution operations \cite{varior2016gated} would be particularly relevant. 

As noted above, we can potentially use a localized feature representation approach to mitigate issues caused by background clutter and noise. In WHOS, another hand-crafted feature representation algorithm, a simple background filter was used to address this issue. Specifically, features were weighted according to their distance to the center. GOG also has the similar trick by weighting patches based on the distance to the center line. While issues like occlusion can certainly create problems, this assumption is not entirely unreasonable. An immediate idea to improve this strategy would be to learn camera-specific background distributions \cite{gou2016person}. At its core, the fundamental idea of the strategy described here and in the previous paragraph is to focus on the ``person'' part of the image as much as possible; we can do this in a much more sophisticated fashion using recent advances in image segmentation \cite{long2015fully,lin2017exploring}, learning the image representation while simultaneously segmenting out the background.

In the context of multi-shot or video-based re-id, we have much more information than a single image for each person, and this can be in the form of a set of images or a video sequence. In addition to the spatial aspect, we can exploit the temporal dimension as well to obtain better feature representations. For instance, we can borrow ideas from the C3D network of Tran \etal \cite{tran2015learning} to learn \textbf{spatio-temporal feature representations} \cite{wang2016temporal} for each available video sequence, following which existing frameworks can be employed to learn discriminative distance metrics. An immediate follow-up to this idea would be the concept of spatio-temporal Siamese networks. While existing Siamese network frameworks learn to tell pairs of images apart, we can extend them, in conjunction with C3D-like networks, to tell pairs of video sequences apart. In multi-shot ranking, we demonstrated that using a custom ranking method gives much better performance compared to using the feature averaging scheme. In practical re-id applications, an image sequence of a person will typically undergo several variations such as background clutter, occlusion, and illumination variations. Developing custom multi-shot ranking algorithms that take this data variance into account will give better performance. Another promising future research direction in this context would be to integrate multi-shot ranking with metric learning. While most existing methods treat these two topics separately, developing a unified metric learning and multi-shot ranking framework that exploits the several aspects of multi-shot data can potentially lead to further performance gains. 

While existing re-id datasets may not be large enough to use recent advances in feature learning, we can use smart augmentation strategies to generate a large number of synthetic images. While common strategies such as random 2-D translations can be readily applied \cite{ahmed2015improved}, we can use sophisticated methods such as CycleGAN \cite{zhu2017unpaired} and LSRO \cite{zheng2017unlabeled} to generate realisitic and meaningful images. For instance, starting from a base person image, we can generate images with different attributes, such as with and without sunglasses, with and without a backpack, and so on, each with a different branch of CycleGAN \cite{gong2017learning}. Furthermore, we can combine these attributes to generate fused images-- for instance, an image of a person with glasses wearing a backpack. Such strategies can be used to generate meaningfully augmented datasets with large number of \textbf{diverse} images, which can in turn help fuel development of new deep learning approaches for re-id, some of which have been discussed in the previous paragraphs. 

Among the various attributes we as humans use to re-identify people, \textbf{walking patterns} or gait are critical \cite{cutting1977recognizing}. When we know who we are looking for, we can pick her/him up far away in a large crowd just by looking at the way s/he walks. Clearly, and intuitively, this information must be exploited by re-id algorithms. While there has been some work along these lines in the past \cite{gou2016person,wang2014person,liu2015enhancing}, much more work needs to be done, specifically in using such dynamic gait-based models in conjunction with appearance modeling via feature learning strategies. 

An interesting but hardly addressed line of research in re-id is the use of \textbf{camera calibration} information. Typically, when one is constructing a new dataset, calibration information for the source cameras is easily obtainable. Such information can prove very handy in reducing the search space of candidates in the gallery. For instance, in a real-world application of re-id \cite{camps2016from, li2014real, assari2016human}, we can use calibration information to estimate motion patterns of people. This can be used to filter candidates that only move in a certain direction, if the operator of the system is sufficiently confident about the trajectory of the person of interest. Furthermore, we can also use calibration information to estimate the ground plane of the scene, thereby helping estimate the height of the candidates seen in the gallery. This information can also be used as a filter to reduce the search space in the gallery. 

Another very interesting aspect that is missing in existing re-id algorithms is the notion of \textbf{context}:  when, where, and with whom is/was the person of interest moving? It is not unreasonable to assume that people often walk in groups, and this information can provide a strong prior to reduce the search space of gallery candidates. For instance, tracking groups of people can help model behavioral trends for the person of interest, providing a rich visual sense of context that can be used to perform re-id. Zheng \etal \cite{zheng2009associating} used this notion of context to perform person matching in a multi-camera setup, and we believe this  is a promising direction to pursue given the availability of several multi-camera re-id datasets.

Other promising directions for future research include using \textbf{multi-modal data} to alleviate potential problems with appearance feature learning. For instance, existing person re-id algorithms would fail in dark rooms or in cases where people wear similar clothes, for instance, in a laboratory or factory setting that mandates a dress code.  In such scenarios, we could use depth information to estimate gait to perform re-id \cite{Haque_2016_CVPR}. Furthermore, much recent work has focused on learning rich visual representations using RGB-D data \cite{song2015sun}, and this can be readily applied to the re-id problem. 

Finally, we conclude with some thoughts on the current state of performance evaluation of re-id algorithms. We believe that researchers should take a broader view of how algorithms perform and not just look at raw rank-1 or mAP numbers. Specifically, as noted in Camps \etal \cite{camps2016from}, a re-id algorithm is only a small, but critical, part of a larger, fully automated system that tracks people in multi-camera networks. To this end, we should focus on creating datasets that accurately mimic scenarios where such systems would be employed. While the airport dataset we propose in this paper is a step towards this direction, much more work needs to be done to achieve practical, realistic evaluation mechanisms for re-id algorithms. For instance, a re-id application may stem from a crime detection/prevention perspective where the perpetrator, or the person of interest, may re-appear after several days or months, and in an entirely new appearance. Clearly, this shows we need datasets that have richer ``temporal'' aspects than those used in this paper. Specifically, datasets that have multiple re-appearances of people, ideally spanning across days or months, and appearances, will help in developing evaluation metrics that shed light on practical, real-world usability of re-id algorithms. A very early idea to this end is proposed in Karanam \etal \cite{karanam2017rank} where the notion of ``rank persistence'' is used to evaluate re-id algorithms. The motivation for this approach is from the crime detection application discussed above; since we do not know when the perpetrator re-appears, the system continuously searches and ranks observed candidates. Once the person of interest in observed, we would like her/him to \textbf{stay} in the top-k rank list for \textbf{as long as possible}. We contend that researchers should focus on developing, and evaluating, re-id algorithms with this notion of persistence of the person of interest over time, resulting in algorithms that are meaningful from a real-world perspective. 

\bibliographystyle{IEEEtran}
\bibliography{references_bib}

\vspace{-4em}
\begin{IEEEbiography}
[{\includegraphics[width=1in,height=1.25in,clip,keepaspectratio]{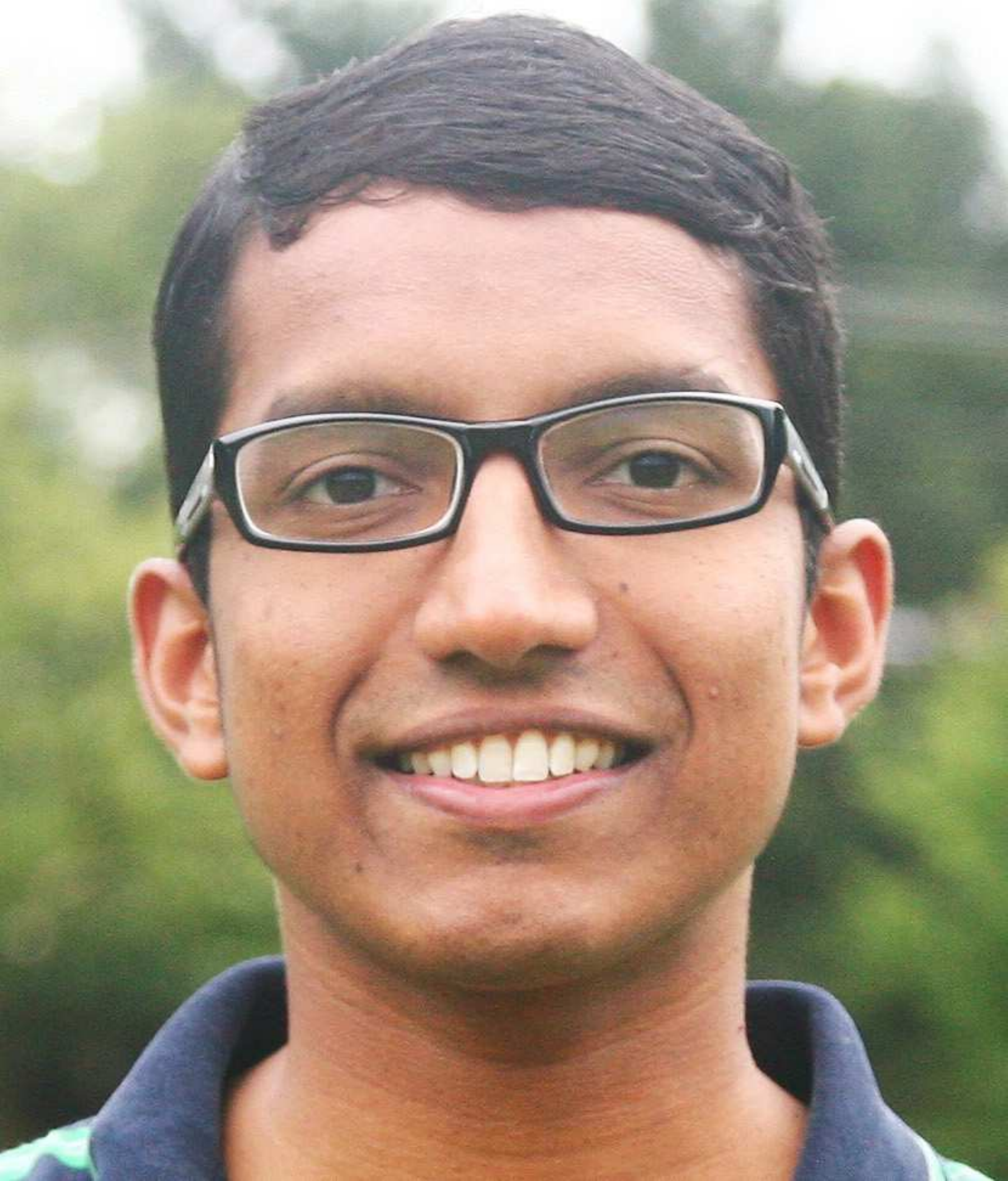}}]{Srikrishna Karanam}
Srikrishna Karanam is a Research Scientist in the Vision Technologies and Solutions group at Siemens Corporate Technology, Princeton, NJ. He has a Ph.D.~degree in Computer \& Systems Engineering from Rensselaer Polytechnic Institute. His research interests include computer vision and machine learning with focus on all aspects of image indexing, search, and retrieval for object recognition applications.
\end{IEEEbiography}

\vspace{-4em}
\begin{IEEEbiography}[{\includegraphics[width=1in,height=1.25in,clip,keepaspectratio]{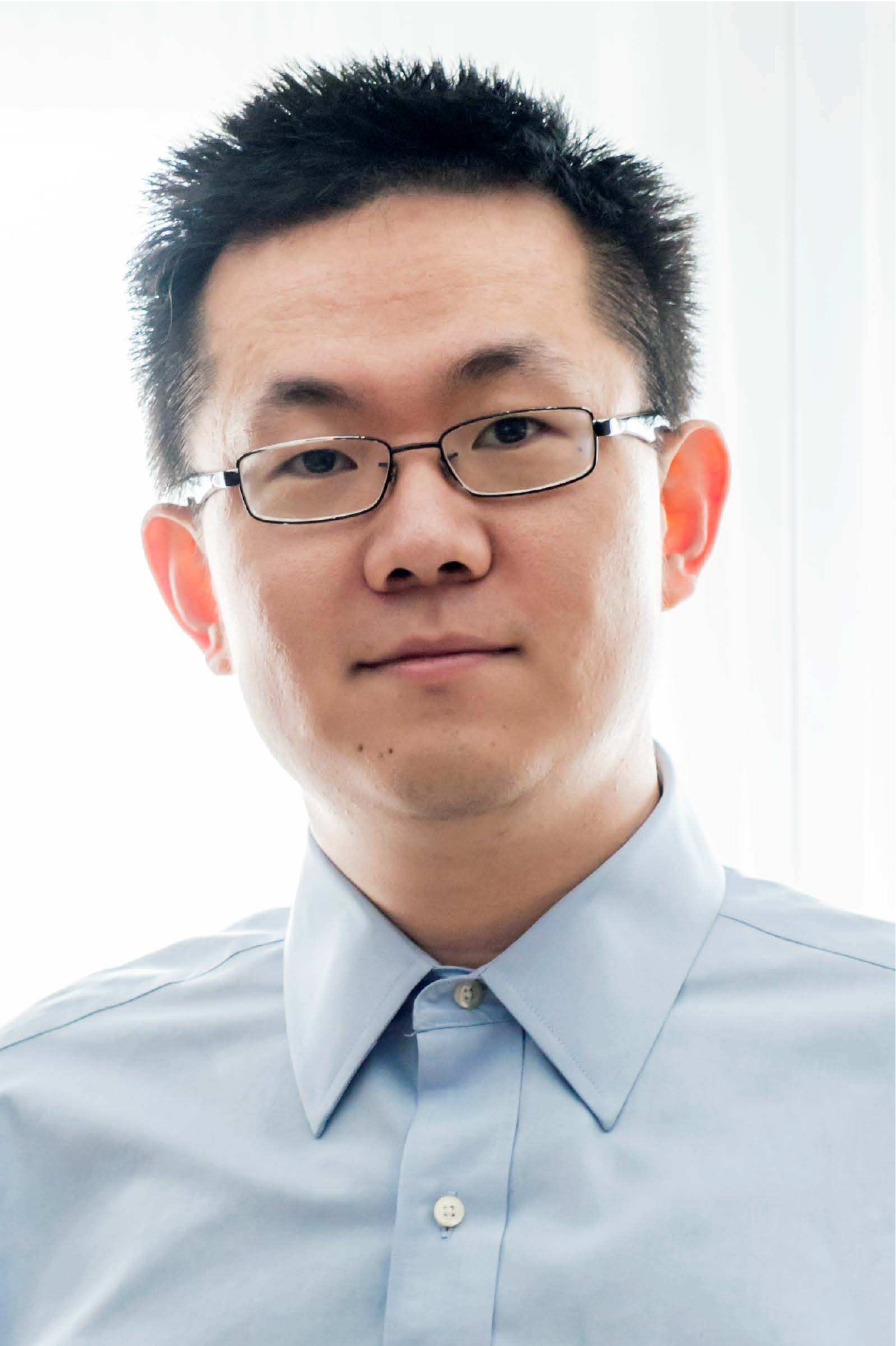}}]{Mengran Guo}
Mengran Gou is currently a Ph.D.~candidate in the Department of Electrical and Computer Engineering at Northeastern University. He received an M.S.~degree from the Pennsylvania State University and B.Eng.~degree from Harbin Institute of Technology in China. His research interests are person re-identification and activity recognition.
\end{IEEEbiography}
\vspace{-1.2cm}

\begin{IEEEbiography}[{\includegraphics[width=1in,height=1.25in,clip,keepaspectratio]{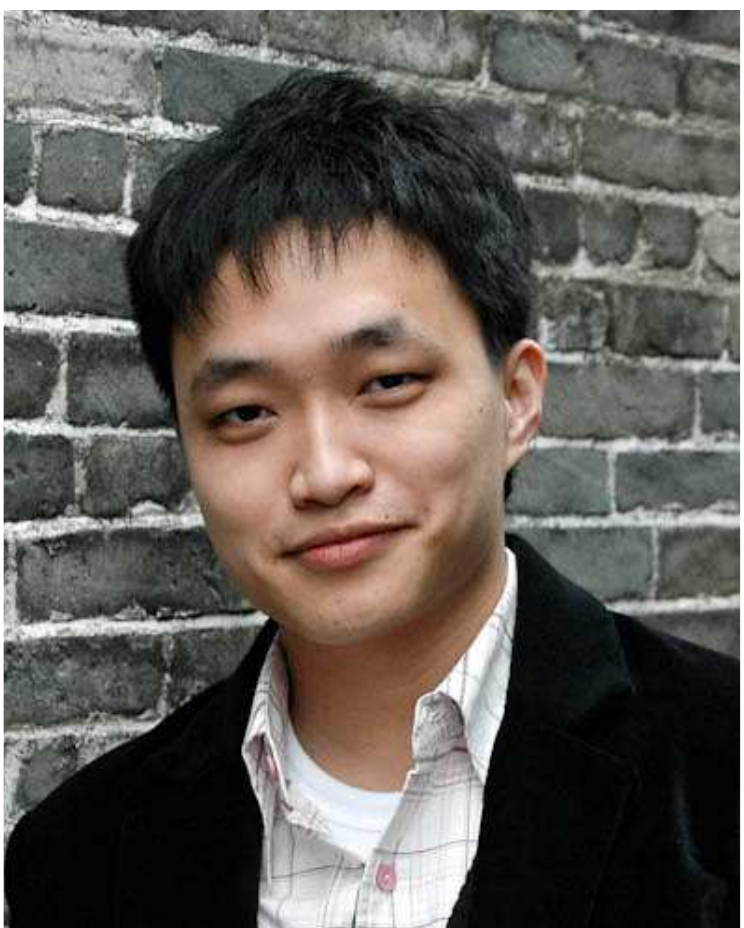}}]{Ziyan Wu}
Ziyan Wu received a Ph.D. degree in Computer and Systems Engineering from Rensselaer Polytechnic Institute in 2014. He has B.S. and M.S. degrees in Engineering from Beihang University. He joined Siemens Corporate Research as a Research Scientist in 2014. His current research interests include 3D object recognition and autonomous perception.
\end{IEEEbiography}
\vspace{-1.2cm}

\begin{IEEEbiography}[{\includegraphics[width=1in,height=1.25in,clip,keepaspectratio]{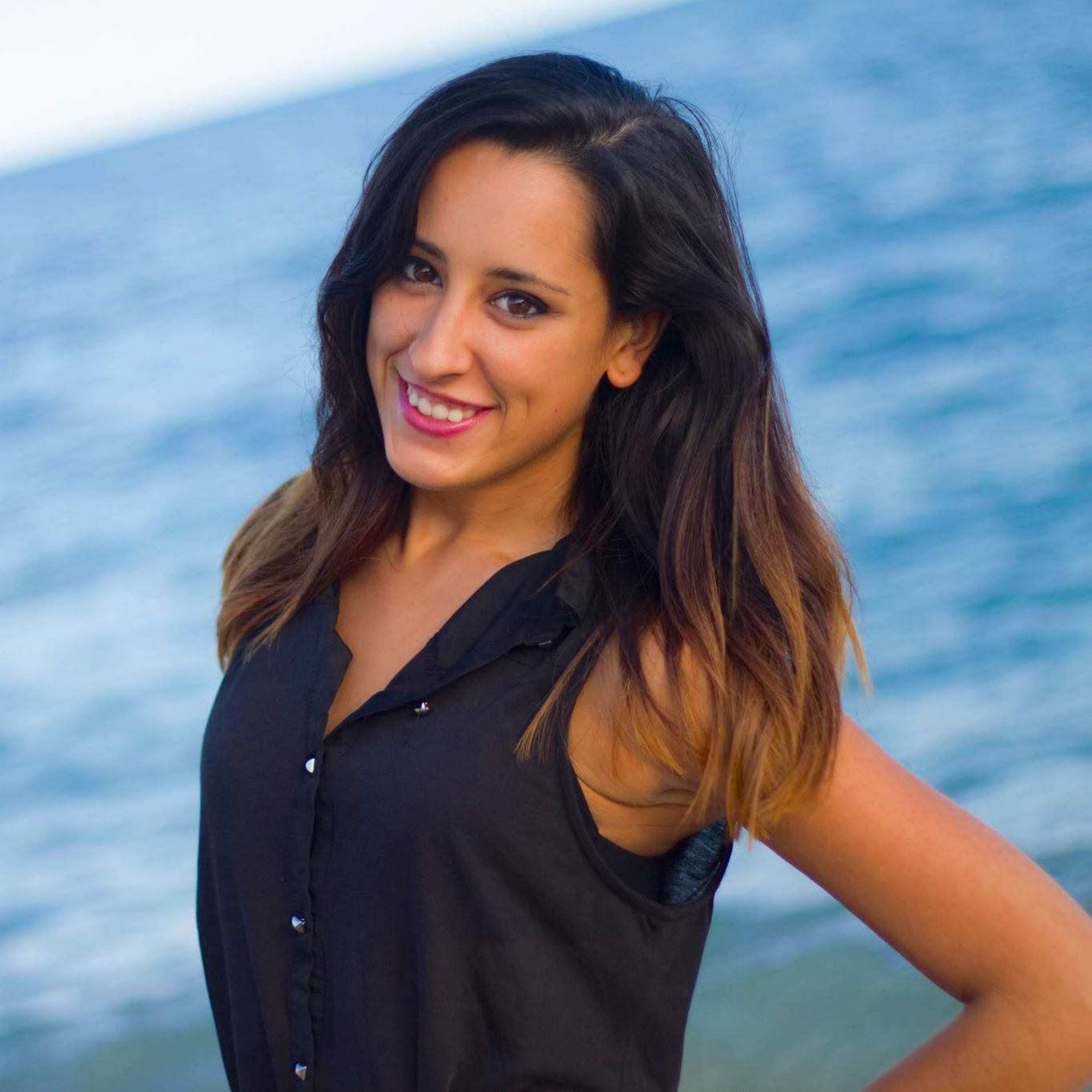}}]{Angels Rates-Borras}
Angels Rates-Borras received her B.S. degree in Electrical Engineering from Universitat Politecnica de Catalunya, Barcelona, Spain. She is currently a Ph.D. student in the Department of Electrical and Computer Engineering at Northeastern University, Boston. Her research interests are mainly focused on person re-identification, scene understanding, video analysis and activity recognition.
\end{IEEEbiography}
\vspace{-1.2cm}

\begin{IEEEbiography}[{\includegraphics[width=1in,height=1.25in,clip,keepaspectratio]{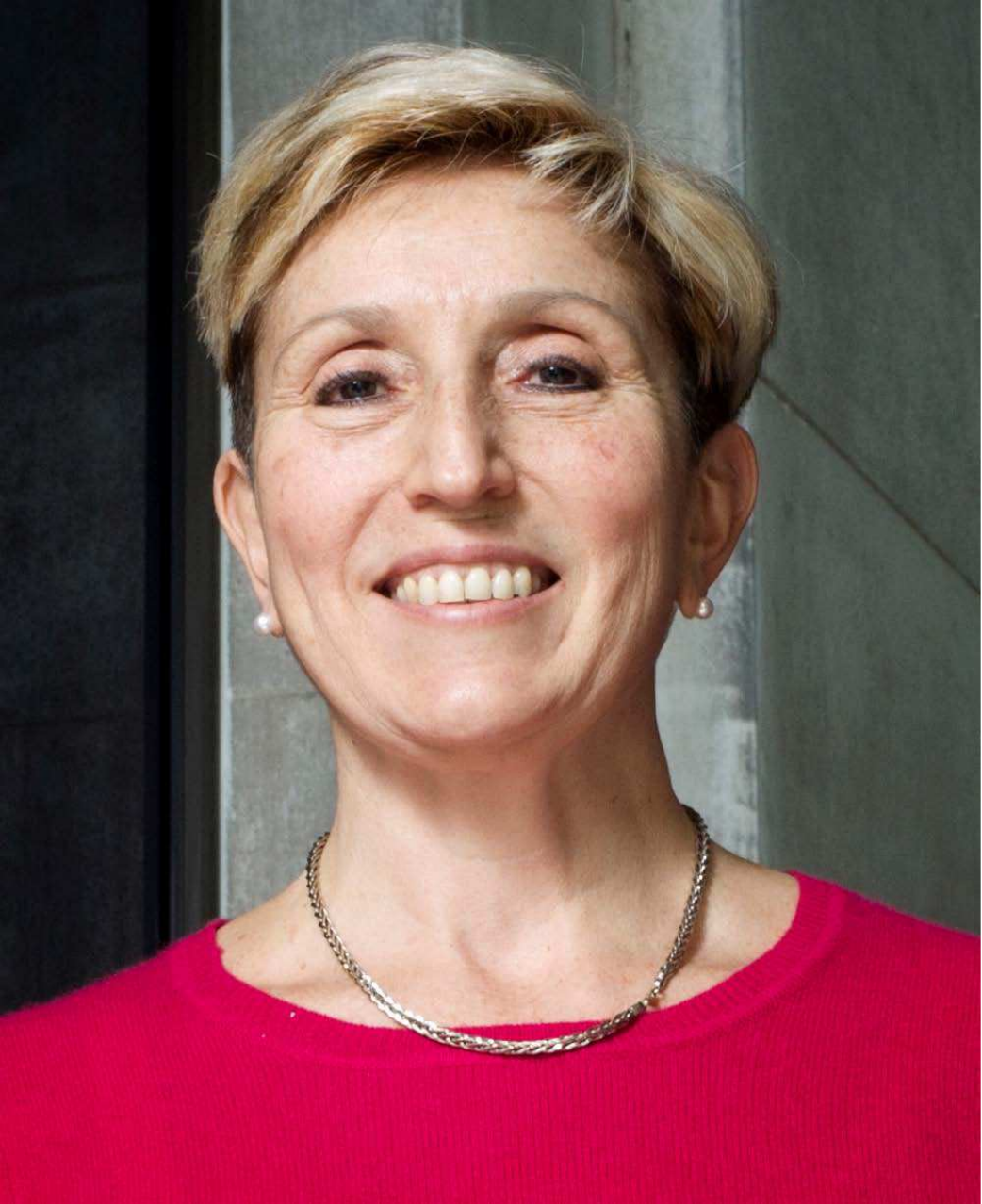}}]{Octavia Camps}
Octavia Camps received B.S.~degrees in computer science in 1981 and in electrical engineering in 1984, from the Universidad de la Republica (Montevideo, Uruguay), and M.S.~and Ph.D.~degrees in electrical engineering in 1987 and 1992, from the University of Washington, respectively. Since 2006 she is a Professor in the Electrical and Computer Engineering Department at Northeastern University. From 1991 to 2006 she was a faculty member at the departments of Electrical Engineering and Computer Science and Engineering at The Pennsylvania State University. In 2000, she was a visiting faculty at the California Institute of Technology and at the University of Southern California and in 2013 she was a visiting faculty at the Computer Science Department at Boston University.  Her main research interests include dynamics-based computer vision, image processing, and machine learning.  She is a member of IEEE.
\end{IEEEbiography}
\vspace{-1.2cm}

\begin{IEEEbiography}[{\includegraphics[width=1in,height=1.25in,clip,keepaspectratio]{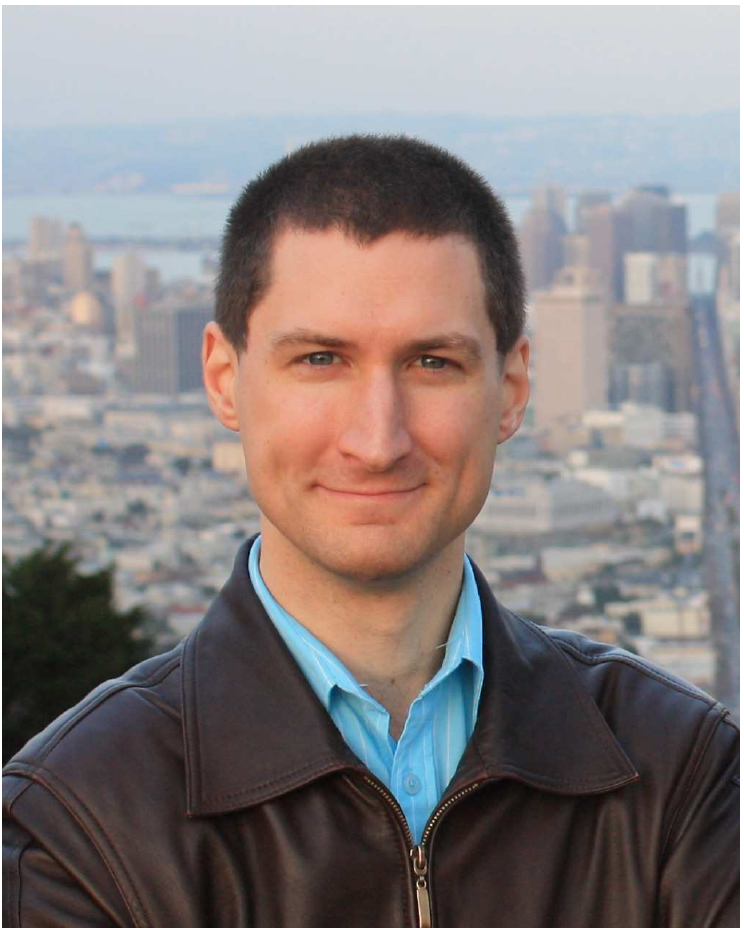}}]{Richard J. Radke}
Richard J.~Radke joined the Electrical, Computer, and Systems Engineering department at Rensselaer Polytechnic Institute in 2001, where he is now a Full Professor. He has B.A.~and M.A.~degrees in computational and applied mathematics from Rice University, and M.A.~and Ph.D.~degrees in electrical engineering from Princeton University.  His current research interests involve computer vision problems related to human-scale, occupant-aware environments, such as person tracking and re-identification with cameras and range sensors.  Dr.~Radke is affiliated with the NSF Engineering Research Center for Lighting Enabled Service and Applications (LESA), the DHS Center of Excellence on Explosives Detection, Mitigation and Response (ALERT), and Rensselaer's Experimental Media and Performing Arts Center (EMPAC). He received an NSF CAREER award in March 2003 and was a member of the 2007 DARPA Computer Science Study Group.  Dr.~Radke is a Senior Member of the IEEE and a Senior Area Editor of \emph{IEEE Transactions on Image Processing}.  His textbook \emph{Computer Vision for Visual Effects} was published by Cambridge University Press in 2012.
\end{IEEEbiography}
\vspace{-1.2cm}

\end{document}